\pdfoutput=1
\RequirePackage{fix-cm}
\documentclass[10pt,twocolumn,letterpaper]{article}

\usepackage[T1]{fontenc}
\usepackage{times}
\usepackage[letterpaper,hmargin=0.63in,top=0.78in,bottom=0.85in,footskip=26pt]{geometry}
\usepackage[hyphens]{url}
\usepackage[dvipsnames]{xcolor}
\usepackage[numbers,sort&compress]{natbib}
\usepackage{graphicx}
\usepackage{amsmath}
\usepackage{amssymb}
\usepackage{booktabs}
\usepackage{multirow}
\usepackage{colortbl}
\usepackage{xspace}
\usepackage{tikz}
\usetikzlibrary{positioning,arrows.meta,shapes.geometric,fit,calc}

\usepackage{microtype}
\usepackage{amsthm,bm}
\usepackage{algorithm,algpseudocode}
\usepackage{stfloats,placeins}
\usepackage[font=small,labelfont=bf,labelsep=period]{caption}
\usepackage{titlesec}
\titleformat{\section}{\large\bfseries}{\thesection}{0.6em}{}
\titleformat{\subsection}{\normalsize\bfseries}{\thesubsection}{0.6em}{}
\titlespacing*{\section}{0pt}{2.1ex plus .5ex minus .3ex}{.9ex}
\titlespacing*{\subsection}{0pt}{1.6ex plus .4ex minus .2ex}{.6ex}
\titlespacing*{\paragraph}{0pt}{1.3ex plus .3ex minus .2ex}{.7em}
\newtheorem{proposition}{Proposition}
\newtheorem{theorem}{Theorem}
\newcommand{\eg}{\emph{e.g}.\xspace}
\newcommand{\ie}{\emph{i.e}.\xspace}
\newcommand{\vs}{\emph{vs}.\xspace}
\definecolor{linkblue}{RGB}{30,75,110}
\usepackage[pagebackref,breaklinks,colorlinks,allcolors=linkblue]{hyperref}

\newcommand{\papertitle}{MOVER: Planning Beyond Motion}
\title{\papertitle}
\author{Xiangyu Li\textsuperscript{1}, Yu Jiang\textsuperscript{1},
  Tianyi Wang\textsuperscript{1}, Liang He\textsuperscript{2},
  Qixing Huang\textsuperscript{1}, Zhaomiao Guo\textsuperscript{1,\textdagger}}
\date{}
\hypersetup{pdftitle={\papertitle},
  pdfauthor={Xiangyu Li, Yu Jiang, Tianyi Wang, Liang He, Qixing Huang, Zhaomiao Guo}}

\makeatletter
\newcommand{\preprinttitle}{%
  {\LARGE\bfseries \@title\par}
  \vspace{0.75em}
  {\large \@author\par}
  \vspace{0.35em}
  {\normalsize \textsuperscript{1}University of Texas at Austin\qquad
    \textsuperscript{2}ETH Zurich\par}
  \vspace{0.25em}
  {\small \textsuperscript{\textdagger}Corresponding author: Zhaomiao Guo
    (\href{mailto:zguo@utexas.edu}{\nolinkurl{zguo@utexas.edu}})\par}}
\makeatother
\renewenvironment{abstract}
  {\par\small\noindent\textbf{Abstract}\par\smallskip\noindent\ignorespaces}
  {\par}
\newcommand{\maketitlesupplementary}{%
  \twocolumn[{\centering
    {\Large\bfseries \papertitle\par}\vspace{0.45em}
    {\large Supplementary Material\par}\vspace{1.4em}}]}

\begin{document}
\twocolumn[{\centering
  \preprinttitle
  \vspace{1.4em}
  \begin{minipage}{0.92\textwidth}
    \begin{abstract}
Residual world models forecast a driving scene through its temporal change, so feature components that stay constant over time vanish from the residual pathway.
We ask whether evidence with weak temporal change, such as a waiting pedestrian or a stopped vehicle, can be used more effectively for planning, and introduce MOVER, a complementary event-routing mechanism.
A training-time teacher measures how each spatially grounded event query responds when a past frame is removed from shared history and contrasts that sensitivity with the event's pooled temporal residual, prioritizing events that depend on history but receive weak residual evidence.
A student predicts the resulting priorities from unperturbed history, so the teacher adds no inference cost.
Under fixed total routing mass, one distribution scales a low-rank future bird's-eye-view update and biases trajectory-conditioned event reading.
On nuScenes, MOVER reduces ResWorld's UniAD-style average L2 from $0.65$ to $0.55\,\mathrm{m}$ and its collision rate from $0.23\%$ to $0.12\%$ without direct planner ego status ($0.50\,\mathrm{m}$ and $0.09\%$ with it), adding $4.1\,\mathrm{ms}$ per frame.
Matched controls show that the complementary teacher outperforms either signal alone and that the allocation acts mainly during training.
Adapted to BridgeDrive--LEAD, MOVER reaches a $96.52\pm0.65$ driving score over three seeds on the 220-route Bench2Drive closed-loop benchmark, exceeding matched controls in each paired seed by a margin within seed variation.
\end{abstract}

  \end{minipage}\par
  \vspace{1.2em}}]
\section{Introduction}
\label{sec:introduction}

\begin{figure*}[t]
	\centering
	\includegraphics[width=0.96\textwidth]{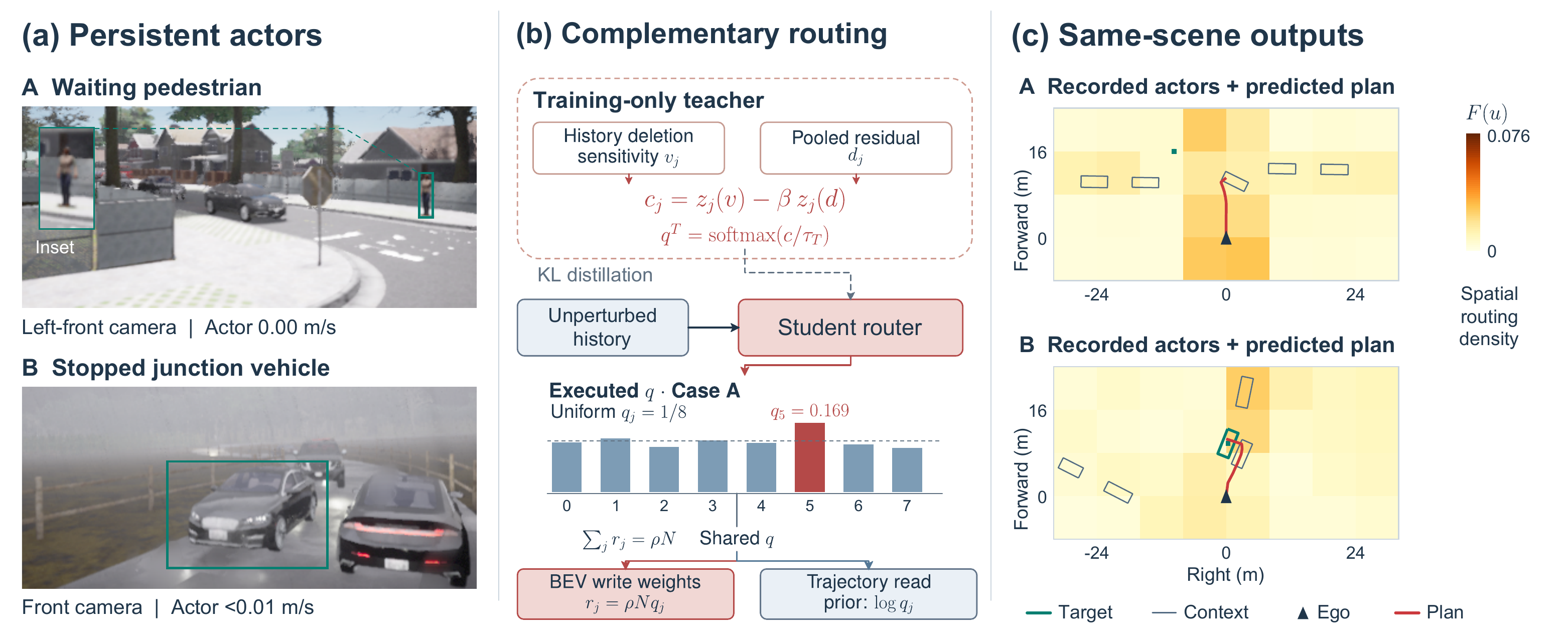}
	\caption{
		\textbf{Persistent traffic context and complementary routing.}
		(a) Recorded CARLA examples show a waiting pedestrian and a stopped congested vehicle.
		(b) A training-only teacher supervises a shared allocation for BEV writing and trajectory reading; all eight bars use example A.
		(c) The same observations show spatial routing density $F(u)=\sum_jq_jm_j(u)$ and predicted plans in local nominal BEV coordinates; green: target actor; blue: context actors; red: plan; density describes allocation rather than object importance.
	}
	\label{fig:novelty}
\end{figure*}

A waiting pedestrian can constrain an upcoming turn, and a stopped vehicle can block a junction before either actor moves (Fig.~\ref{fig:novelty}a).
Planning-oriented systems organize perception and prediction around the driving task~\cite{hu2023uniad,jiang2023vad}, and residual world models forecast the scene through its temporal change~\cite{zhang2026resworld}.
Temporal differencing removes time-constant feature components, so evidence that persists unchanged contributes little to the residual pathway, although the full predictor can still retain current-scene content.
For residual world models, we ask: \emph{can supplementary evidence allocation improve planning when pooled temporal change is weak?}

Recent driving world models predict the future through different representations.
LAW~\cite{li2025law} learns predictive scene features conditioned on ego trajectories, DriveLaW~\cite{xia2026drivelaw} connects video-generation latents directly to a trajectory planner, and ResWorld~\cite{zhang2026resworld} separates dynamic prediction from persistent scene content by combining predicted residual features with current bird's-eye-view (BEV) features.
ResWorld also identifies pedestrians and parked vehicles as difficult cases for temporal residuals and uses a prior trajectory branch to handle them.
The open question is whether a model can identify retained evidence whose additional use improves planning.

MOVER prioritizes supplementary evidence by \emph{complementarity}.
A weak pooled residual may reflect little net change, cancellation within the event mask, or uninformative background.
History sensitivity can be high for events that already carry strong dynamic evidence.
Combining the two signals gives an event a high complementary score when its representation depends on historical evidence and its pooled residual remains weak (Fig.~\ref{fig:novelty}).
Neither signal suffices alone: a residual-only teacher trails even a router trained without a teacher, and a sensitivity-only teacher trails the combined score (Table~\ref{tab:exp_routes}).
The score is model-relative and ranks learned spatial events rather than objects, while stationarity alone does not make an event complementary.

We introduce \textbf{MOVER}, a complementary event-routing mechanism for a residual BEV world model.
Each event is a learned spatial summary rather than a detected object or predefined semantic category.
Spatially biased soft masks extract current event content and pool corresponding regions across ego-aligned history.
During training, we remove one past frame from the history-attention readout while preserving the current observation.
For event $j$, let $v_j$ measure the resulting squared change in its contextual representation, and let $d_j$ denote the squared norm of its pooled temporal residual.
The complementary score is
\begin{equation}
	c_j = z_j(v) - \beta z_j(d),
	\label{eq:intro_complementarity}
\end{equation}
where $z_j(\cdot)$ standardizes each signal across events within a scene and $\beta \geq 0$ weights residual evidence.
The scores define a teacher distribution, which we distill into a router that runs on the unperturbed history, so inference preserves the learned priorities without the extra readout.

MOVER uses the same allocation at the two interfaces through which events influence the plan.
For $N$ events with student probabilities $q_j^S$, writing weights are $r_j=\rho Nq_j^S$, where $\rho$ sets their mean and fixes the total routing mass to $\rho N$.
These weights scale a low-rank correction to future BEV features, and the same probabilities act as a prior for trajectory-conditioned attention over history-enhanced events, linking the evidence written into the predicted scene to the evidence read during trajectory refinement.

Therefore, MOVER (i) proposes a \textbf{complementary scoring rule} from history-deletion sensitivity and pooled residual evidence that requires no event-importance labels, (ii) uses one \textbf{fixed-sum route distribution} for BEV writing and as a trajectory-attention prior, and (iii) \textbf{locates the gain} with matched controls and fixed-checkpoint route swaps, showing that the allocation acts mainly during training.
A supporting analysis in the supplementary material bounds how teacher--student route mismatch propagates through both paths of a fixed model, which is a stability statement, not a guarantee of planning quality.

We evaluate MOVER on nuScenes~\cite{caesar2020nuscenes} under the VAD- and UniAD-style planning protocols.
Without direct planner ego status, MOVER attains average L2 errors of $0.28\,\mathrm{m}$ and $0.55\,\mathrm{m}$ with collision rates of $0.04\%$ and $0.12\%$, against $0.35\,\mathrm{m}$, $0.65\,\mathrm{m}$, $0.07\%$, and $0.23\%$ for ResWorld.
With direct planner ego status, the respective L2 errors are $0.24\,\mathrm{m}$ and $0.50\,\mathrm{m}$, and collision rates are $0.03\%$ and $0.09\%$.
We also evaluate the routing interface in closed-loop driving on Bench2Drive~\cite{jia2024bench2drive}, using BridgeDrive--LEAD~\cite{liu2026bridgedrive,nguyen2026lead} as the base planner: MOVER exceeds matched controls in each of three paired seeds, by $0.60$ driving-score points on average over uniform routing, a margin within seed variation.

\section{Related Work}
\label{sec:related_work}

\subsection{Planning-Aware Scene Representations}

Driving planners encode action-relevant structure with shared spatial-temporal features and queries~\cite{hu2022stp3,hu2023uniad,wang2025hcomc}, or with compact agent and map representations~\cite{jiang2023vad,sun2025sparsedrive}.
SSR~\cite{li2025ssr} selects navigation-guided scene tokens; SpaceDrive~\cite{li2026spacedrive} grounds vision-language planning in spatial features; SafeDrive~\cite{kim2026safedrive} evaluates collision and road compliance in trajectory-conditioned sparse worlds.
Generative planners sample action hypotheses~\cite{zheng2024genad,liao2025diffusiondrive,sun2026sparsedrivev2}.
These approaches choose what reaches the planner, while MOVER allocates supplementary evidence to events that depend on history while receiving weak residual evidence, without event-importance labels.

\subsection{World Models and Evidence Allocation}

Driving world models forecast occupancy~\cite{zheng2024occworld,yang2025driveoccworld,wang2026vehdyn}, generate controllable video~\cite{wang2024drivedreamer,wang2024drivewm,cheng2026m4world}, or use future images to supervise policies~\cite{li2026drivevlaw0}.
LAW~\cite{li2025law} and DriveLaW~\cite{xia2026drivelaw} pass predictive latents or video-generation features to a planner.
Other methods train intermediate world representations, unified vision--motion spaces, scene--action hypotheses, and rewards from imagined outcomes~\cite{gui2026worlddrive,xia2026reworld,liu2026mmfuture}.
ResWorld~\cite{zhang2026resworld} combines residual predictions with current BEV features and acknowledges difficult pedestrian and parked-vehicle cases; DynFlowDrive~\cite{liu2026dynflowdrive} uses rectified flow for action-conditioned latent transitions; Drive-HWM~\cite{fan2026drivehwm} separates slow dynamics from fast action generation.
Because a weak pooled residual can also result from cancellation or uninformative background, MOVER combines residual magnitude with history sensitivity to guide supplementary BEV writing and event reading, without treating either signal as ground-truth planning relevance.

\subsection{Adaptive Tokens, Distillation, and Evaluation}

TokenLearner~\cite{ryoo2021tokenlearner} forms learned spatial summaries, and Slot Attention~\cite{locatello2020slot} forms object-centric slots.
Token pruning and merging reduce computation~\cite{rao2021dynamicvit,bolya2023tome}, including in driving-specific selectors~\cite{sha2026stprune,yang2026mvpruner}.
MOVER retains every event and redistributes a fixed routing mass over their influence on scene writing and trajectory reading.
Counterfactual relevance methods infer object importance from synthetic changes in driving decisions~\cite{lusvarghi2026relevance,zhang2026counterfactual}.
MOVER deletes one past frame from attention while retaining the current observation, akin to occlusion sensitivity~\cite{zeiler2014visualizing} applied along time, and uses the change in event context as a model-relative training target.
Teacher--student distillation~\cite{hinton2015distill,chen2020cheating}, perturbed-view consistency~\cite{tarvainen2017mean,assran2023ijepa}, and entropy/KL-regularized allocation~\cite{geist2019regularized} provide the training tools; no event-importance annotations are required.
On nuScenes~\cite{caesar2020nuscenes}, displacement error and collision avoidance can respond differently to perception changes~\cite{zhai2023rethinking}, and direct ego status can dominate open-loop scores~\cite{li2024egostatus}.
NAVSIM~\cite{dauner2024navsim}, Bench2Drive~\cite{jia2024bench2drive}, and cross-city~\cite{naeinian2026crosscity} protocols test different properties, while method rankings can invert across benchmarks~\cite{wang2026crossbenchmark}.

\section{Methodology}
\label{sec:method}

\begin{figure*}[t]
	\centering
	\includegraphics[width=0.94\textwidth]{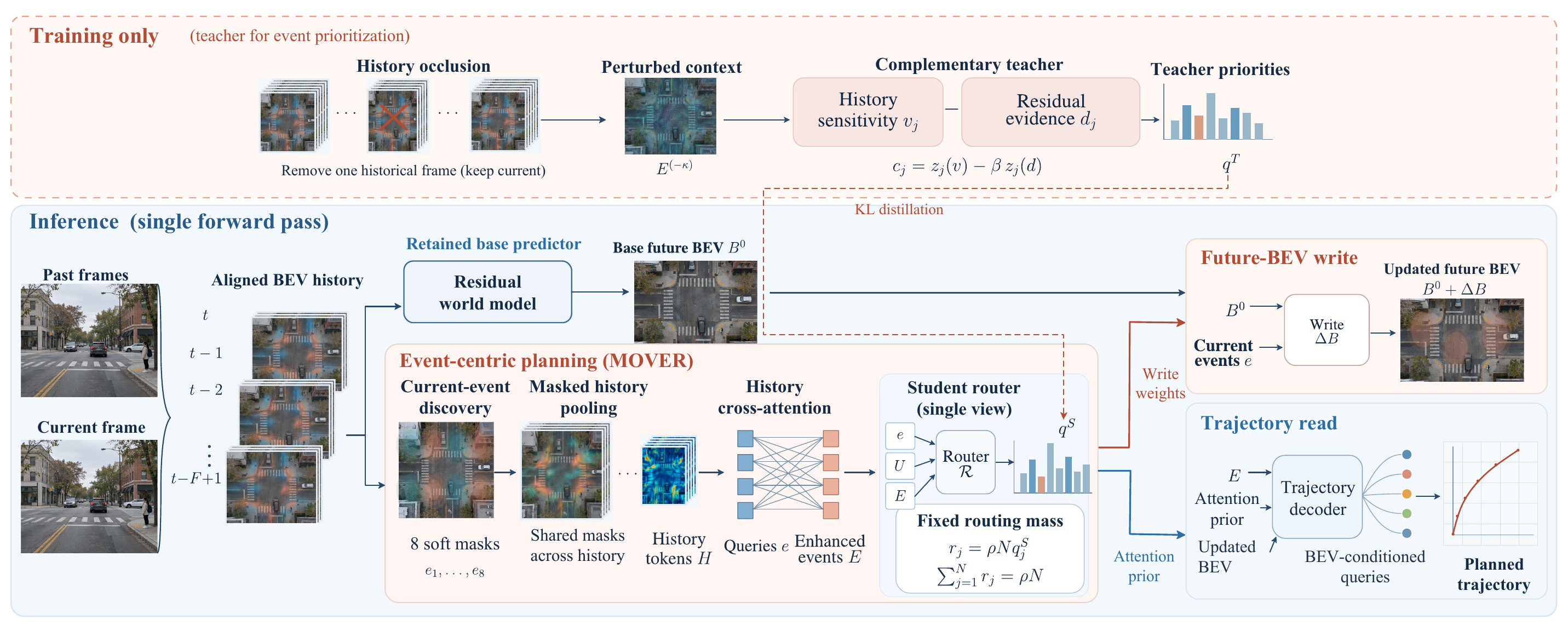}
	\caption{
		\textbf{Overview of MOVER.}
		A residual driving world model predicts future BEV features from ego-aligned history, and MOVER adds a complementary event-routing branch.
		We discover current-frame events once, pool the corresponding regions across aligned history, and enrich them with shared history cross-attention.
		During training, a history-occlusion teacher produces complementary priorities from history sensitivity and residual evidence; a single-view student router distills these priorities.
		At inference, the learned allocation writes a low-rank update to the future BEV representation and provides a prior for trajectory-conditioned event reading.
		The base predictor supplies the future BEV baseline while MOVER routes supplementary evidence.
	}
	\label{fig:framework}
\end{figure*}

Built on a residual BEV world model, MOVER adds a complementary event-routing mechanism: an event receives additional emphasis when removing history changes its representation while the residual pathway supplies weak local evidence.
Waiting pedestrians, stopped vehicles, and partially occluded actors motivate this setting, but low residual magnitude alone does not establish their planning relevance.
A training-time teacher ranks events by this \emph{complementarity}, and a single-view router distilled from the teacher executes the allocation at inference (Fig.~\ref{fig:framework}; Supplementary Fig.~\ref{fig:module} details the routing module).

\subsection{Problem Formulation}
\label{subsec:problem_formulation}

Let $F \ge 2$ denote the number of ego-aligned BEV frames, indexed by $f=0,\dots,F-1$, where $f=0$ is the current frame and $f=1$ is the nearest history frame.
Each frame is represented by $X_f \in \mathbb{R}^{M \times C}$, where $M=H_{\mathrm{bev}}W_{\mathrm{bev}}$ is the number of BEV cells and $C$ is the channel dimension.
We omit the sample index unless needed.

The base predictor outputs a future BEV representation $B^0$.
MOVER adds an event-induced correction $\Delta B(q)$ controlled by a routing distribution $q \in \Delta_N$, where
\[
\Delta_N = \left\{ q \in \mathbb{R}^N \,\middle|\, q_j \ge 0,\ \sum_{j=1}^N q_j = 1 \right\},
\]
and $N$ is the number of event tokens.
The final future BEV is $\widehat B(q) = B^0 + \Delta B(q)$.

Let $w_i(\widehat B(q))$ denote the $i$-th trajectory query read from the future BEV, where $i$ indexes all navigation modes and future steps.
Let $d(\cdot)$ be the base trajectory decoder and $C_i(q)$ the route-conditioned event context directly read by the planner.
MOVER predicts a residual correction $g(\cdot,\cdot)$ and outputs the step displacement
\begin{equation}
	\widehat\delta_i(q)
	= d\!\left(w_i(\widehat B(q))\right)
	+ g\!\left(w_i(\widehat B(q)), C_i(q)\right),
	\label{eq:delta_prediction}
\end{equation}
with cumulative position $\widehat p_h(q) = \sum_{t=1}^{h} \widehat\delta_t(q)$.

The non-$\diamond$ planner receives CAN-bus-conditioned BEV features without direct ego-status concatenation; MOVER$\diamond$ also uses ego status in the planner.
Beyond the base predictor's trajectory supervision, training adds a cumulative-position loss and a clearance loss that use future actor annotations; inference uses unperturbed history and no future annotations.

\paragraph{What temporal differencing removes.}
The residual pathway encodes temporal change.
If $\mathsf D$ denotes temporal differencing and $a \in \mathbb{R}^C$ is a time-constant feature, then $\mathsf D(\mathbf 1_F \otimes a) = 0$, where $\mathbf 1_F$ is the all-ones vector and $\otimes$ denotes repetition across frames.
This identity concerns time-constant feature components, not the loss of traffic constraints from the full predictor.
The base model retains current BEV content; MOVER tests an additional route for using it in planning.

\subsection{Current-Event Discovery and Shared History Context}
\label{subsec:event_discovery}

We extract $N$ current-frame event tokens from $X_0$; each event is a learned spatial summary.

For event $j$, let $\pi_j(u)>0$ be a spatial reference prior over cell $u \in \{1,\dots,M\}$, and let $\ell_{ju}$ be a learned content score.
The normalized event mask is
\begin{equation}
	m_{ju} = \frac{\pi_j(u)\exp(\ell_{ju})}
	{\sum_{v=1}^{M}\pi_j(v)\exp(\ell_{jv})}.
	\label{eq:event_mask}
\end{equation}
For the localized configuration, $j=(r,c)$ indexes a row-major $R\times K$ lattice, with normalized cell center $(y_u,x_u)$ and event center $(\bar y_r,\bar x_c)=((r+\tfrac12)/R,(c+\tfrac12)/K)$.
We set
\begin{equation}
\pi_{rc}(u)=\exp\!\left[-\frac12\left(\frac{R(y_u-\bar y_r)}{\sigma}\right)^2
-\frac12\left(\frac{K(x_u-\bar x_c)}{\sigma}\right)^2\right],
\label{eq:spatial_prior}
\end{equation}
using $R=2$, $K=4$, and $\sigma=0.5$ in the final model.
Using these masks, we form a current event token
\begin{equation}
	e_j = \mathcal P_e \!\left(\sum_{u=1}^{M} m_{ju} Y_{0u}\right),
	\qquad
	Y_f = [X_f, P],
	\label{eq:event_token}
\end{equation}
where $P$ is a shared spatial position encoding and $\mathcal P_e$ is a linear projection.
We retain a current local context summary $U_j = \operatorname{LN}(\sum_{u} m_{ju} X_{0u})$.

The same current-event masks are reused across aligned history, preserving spatial correspondence between an event now and the evidence carried at that location in recent frames.
For each frame $f$, we pool the masked content and encode it with a temporal embedding
\begin{equation}
	H_{fj}
	=
	\mathcal E_h\!\left(
		\mathcal P_h\!\left(\sum_{u=1}^{M} m_{ju} Y_{fu}\right)+t_f
	\right),
	\label{eq:history_pool}
\end{equation}
where $\mathcal P_h$ is a history projection, $t_f$ a learned temporal embedding, and $\mathcal E_h$ a light history encoder.
Stacking all $H_{fj}$ yields $FN$ event-history tokens, far fewer than a full-grid temporal branch.
We enrich the current events through shared cross-attention
\begin{equation}
	E = \operatorname{LN}\!\left(e + \operatorname{MHA}(\operatorname{LN}(e), H, H)\right),
	\label{eq:history_context}
\end{equation}
where the $j$-th row $E_j$ is the history-enhanced representation of event $j$.
The query for $E_j$ comes from a spatially grounded current event, while its attention keys and values contain $H_{fk}$ for every frame $f$ and event $k$.

\subsection{Complementary Teacher and Student Router}
\label{subsec:teacher}

\paragraph{History sensitivity.}
We sample one historical frame index $\kappa \in \{1,\dots,F-1\}$ uniformly and remove that frame from the history attention keys and values while keeping the current observation intact.
Let $E^{(-\kappa)}$ denote the resulting perturbed event representation.
We define the event-wise history sensitivity as
\begin{equation}
	v_j = \frac{1}{4}\|E_j - E_j^{(-\kappa)}\|_2^2.
	\label{eq:history_sensitivity}
\end{equation}
A large $v_j$ indicates that the local event query depends on the removed frame's shared historical context; removing frame $\kappa$ removes all $H_{\kappa k}$, not only $H_{\kappa j}$.
The perturbation is thus global: $v_j$ also responds to background and alignment changes shared across events, and sampling one $\kappa$ per step estimates the sensitivity averaged over history frames.
The factor $1/4$ only rescales $v$ and is removed by the standardization in Eq.~\eqref{eq:standardize}.

\paragraph{Residual evidence.}
We use the squared norm of an event's pooled temporal residual as a proxy for the dynamic signal at its spatial support
\begin{equation}
	d_j
	=
	\left\|
	\sum_{u=1}^{M} m_{ju}(X_{0u}-X_{1u})
	\right\|_2^2.
	\label{eq:residual_evidence}
\end{equation}
Pooling precedes the squared norm; $d_j$ therefore measures the net residual vector within the event support, over the most recent frame pair only.

\paragraph{Complementary score.}
Within each sample, we standardize both signals across events
\begin{equation}
	z_j(x)
	=
	\frac{x_j-\bar x}
	{\sqrt{N^{-1}\sum_{k=1}^{N}(x_k-\bar x)^2}+\epsilon},
	\label{eq:standardize}
\end{equation}
with $\bar x = N^{-1}\sum_{k}x_k$.
The teacher complementary score and routing distribution are
\begin{equation}
	c_j = z_j(v) - \beta z_j(d), \qquad
	q^T = \operatorname{softmax}(c/\tau_T),
	\label{eq:comp_score}
\end{equation}
with $\beta \ge 0$ and temperature $\tau_T>0$.
A high score combines dependence on the removed history with a weak pooled residual.
Both signals are model-relative training targets, not measures of planning relevance: content that the current frame alone determines has a small $v_j$ whatever its motion, so the score favors history-dependent evidence with a weak net residual and does not by itself single out stationary actors.

The teacher runs during training.
At inference, a student router receives the complete-history features and performs one unperturbed forward pass
\begin{equation}
	a_j = \mathcal R([e_j, U_j, E_j]),
	\qquad
	q^S = \operatorname{softmax}(a/\tau_S),
	\label{eq:student_router}
\end{equation}
where $\mathcal R$ is a lightweight MLP.
We distill the teacher distribution by minimizing
\begin{equation}
	\mathcal L_{\mathrm{KL}}
	=
	D_{\mathrm{KL}}(\operatorname{sg}(q^T)\,\|\,q^S),
	\label{eq:distill_loss}
\end{equation}
with $\operatorname{sg}(\cdot)$ denoting stop-gradient.

\subsection{One Allocation, Two Uses}
\label{subsec:two_uses}

To control \emph{where} route weight goes while fixing its sum, we convert routing probabilities into exact per-sample route mass
\begin{equation}
	r_j = \rho N q_j^S,
	\qquad
	\sum_{j=1}^{N} r_j = \rho N,
	\label{eq:exact_budget}
\end{equation}
where $\rho \in (0,1]$ is the mean routing mass.
The magnitude of the BEV update also depends on the scatter weights, event features, and learned scale $\eta$.

\paragraph{Route-controlled BEV writing.}
Let $f_j = \mathcal P_B(e_j)$ be the write feature of event $j$, and let $\mathsf F_B \in \mathbb{R}^{N\times C}$ stack all such features.
With spatial scatter weights $\mathsf S \in \mathbb{R}^{M\times N}$ derived from the event masks
\begin{equation}
	S_{uj}
	=
	\frac{m_{ju}}
	{\max\!\left(\sum_{k=1}^{N}m_{ku},\epsilon\right)},
	\qquad
	\Delta B(q)
	=
	\eta\, \mathsf S \operatorname{diag}(r)\mathsf F_B,
	\label{eq:bev_write}
\end{equation}
where $\eta$ is a learnable residual scale initialized small.
MOVER adds a structured residual to the future BEV, and since $\operatorname{rank}(\Delta B(q)) \le N$, the supplementary update is low-rank by construction.

\paragraph{Route-conditioned trajectory reading.}
The same allocation also enters the planner directly.
With $w_i(\widehat B(q))$ the $i$-th waypoint feature read from the updated BEV, we compute event-query attention from content compatibility and a route prior
\begin{equation}
	A_{ij}(q)
	=
	\frac{\exp(\ell^{\mathrm{tr}}_{ij})\,\widetilde q_j^{\gamma}}
	{\sum_{k=1}^{N}\exp(\ell^{\mathrm{tr}}_{ik})\,\widetilde q_k^{\gamma}},
	\qquad
	\ell^{\mathrm{tr}}_{ij}=\frac{Q_iK_j^\top}{\sqrt{d_r}},
	\label{eq:traj_attention}
\end{equation}
with $\widetilde q_j = \max(k_\rho q_j^S,\epsilon_p)$, and form the event context $C_i(q) = \sum_{j} A_{ij}(q)V_j$.
Here $k_\rho q_j^S=r_j/(\sum_k r_k+10^{-8})$ renormalizes the route mass to a probability, so $k_\rho=\rho N/(\rho N+10^{-8})\approx 1$; the floor $\epsilon_p=10^{-6}$ keeps the log-prior finite, and we set $\gamma=1$.
The exponent $\gamma \ge 0$ controls the route prior: at $\gamma=0$ the readout reduces to standard content attention, and at $\gamma>0$ the learned allocation biases the planner toward events the router considers more complementary.
The planner then predicts a bounded residual trajectory correction
\begin{equation}
	g(w_i,C_i)
	=
	\delta_{\max}\tanh\!\big(\mathcal M([w_i,C_i])\big),
	\label{eq:traj_residual}
\end{equation}
where $\delta_{\max}=0.5\,\mathrm m$ bounds each correction coordinate and the output layer of $\mathcal M$ is initialized to zero, so the direct trajectory residual starts at zero.
BEV writing thus changes the future scene and the trajectory queries read from it, whereas event reading adds route-conditioned context to the trajectory decoder.

\subsection{Training Objective and Implementation}
\label{subsec:training_objective}

The complete objective is
\begin{equation}
	\mathcal L
	=
	\mathcal L_{\mathrm{base}}
	+
	\lambda_{\mathrm{KL}}\mathcal L_{\mathrm{KL}}
	+
	\lambda_{\mathrm{pos}}\mathcal L_{\mathrm{pos}}
	+
	\lambda_{\mathrm{clr}}\mathcal L_{\mathrm{clr}},
	\label{eq:full_objective}
\end{equation}
where $\mathcal L_{\mathrm{base}}$ includes the base predictor's original objectives, including trajectory supervision.
The cumulative position loss supervises whole paths rather than isolated offsets
\begin{equation}
	\mathcal L_{\mathrm{pos}}
	=
	\frac{\sum_{b}\sum_{h}
		\omega_h \nu_{bh}\,
		\phi\!\left(\widehat p_{bh}-p^{\mathrm{gt}}_{bh}\right)}
	{\max\!\left(\sum_{b}\sum_{h}\omega_h \nu_{bh},\,1\right)},
	\label{eq:pos_loss}
\end{equation}
with $\nu_{bh}$ a valid-prefix mask, $\omega_h$ a horizon weight, and $\phi(e) = \sqrt{\|e\|_2^2 + 10^{-8}} - 10^{-4}$.
The annotation-based clearance surrogate
\begin{equation}
	\mathcal L_{\mathrm{clr}}
	=
	\frac{1}{n_b}\sum_{b}
	\frac{\sum_{h}\max_{a\in\mathcal A_b}
		\left\{\zeta_{bah}[1-\psi_{bah}(\widehat p_{bh})]_+^2\right\}}
	{\max\!\left(\sum_{h}\nu_{bh},1\right)}
	\label{eq:clearance_loss}
\end{equation}
penalizes overlap with annotated future actors.
Here $n_b$ is batch size, $\mathcal A_b$ is the eligible actor set, $\psi_{bah}$ is maximum normalized separation along the actor's oriented axes after expanding its box by the ego footprint and a $0.2$-m margin, and $\zeta_{bah}$ selects valid actor futures on valid ego prefixes while excluding annotation conflicts with the reference trajectory.
The maximum over an empty eligible-actor set is zero; the exact geometry and gate are given in the supplementary material.

The main configuration uses $N=8$ event tokens, $\tau_T=\tau_S=0.5$, $\beta=1$, $\rho=0.25$, residual-scale initialization $\eta=0.1$, and $\lambda_{\mathrm{KL}}=10^{-3}$.
At inference, MOVER adds five operations (current-frame event discovery, masked history pooling, one event-to-history cross-attention, one low-rank scatter write, and one route-conditioned trajectory read) whose cost scales linearly with BEV size and temporal length for the small fixed $N$ used here (Supplementary Sec.~\ref{sec:suppl_inference_profile}).
The teacher adds one masked cross-attention pass during training only; MOVER uses the same sensor inputs and evaluation protocol as the base predictor.

\setcounter{topnumber}{2}
\setcounter{bottomnumber}{2}
\setcounter{totalnumber}{4}
\setcounter{dbltopnumber}{1}
\renewcommand{\topfraction}{0.88}
\renewcommand{\bottomfraction}{0.70}
\renewcommand{\textfraction}{0.10}
\renewcommand{\floatpagefraction}{0.75}
\renewcommand{\dbltopfraction}{0.85}
\renewcommand{\dblfloatpagefraction}{0.75}
\setlength{\textfloatsep}{9pt plus 2pt minus 2pt}
\setlength{\dbltextfloatsep}{9pt plus 2pt minus 2pt}
\setlength{\floatsep}{7pt plus 2pt minus 2pt}
\setlength{\dblfloatsep}{7pt plus 2pt minus 2pt}
\setlength{\intextsep}{7pt plus 2pt minus 2pt}

\section{Experiments}
\label{sec:experiments}

\subsection{Evaluation Setup}
\label{sec:exp_setup}

\paragraph{Data and prediction task.}
We evaluate on nuScenes~\cite{caesar2020nuscenes}, using the residual BEV predictor of ResWorld~\cite{zhang2026resworld} as the base model.
The planner predicts six two-dimensional displacements at $0.5$-second intervals, covering a $3$-second future; the navigation command selects the evaluated mode.
The validation set contains $6{,}019$ observations; planning and command-conditioned allocation summaries use the $5{,}119$ with a complete future.
Evaluation is open-loop: predictions are compared with recorded futures and are not fed back into the scene.

\paragraph{Temporal aggregation.}
We report both the VAD-style/ST-P3 time-averaged protocol, which averages position errors over all steps up to each horizon, and the UniAD-style endpoint protocol, which uses the error at the horizon; collision rates use the corresponding temporal aggregation of the planning evaluator's collision indicators.
Each reported average is the arithmetic mean of its $1$, $2$, and $3$-second scores, so the average VAD-style L2 weights the six future steps by $(11,11,5,5,2,2)/36$ (Supplementary Eq.~\eqref{eq:exp_prefix_weights}).

\paragraph{Controlled comparisons.}
The component suite keeps initialization, freezing policy, training data, and training schedule fixed while varying the trajectory refiner and its auxiliary objectives. The routing suite keeps model capacity and the refiner fixed while varying the routing strategy, the teacher target, or the auxiliary losses.
Fixed-parameter route diagnostics use a localized model.
nuScenes entries are reported without seed-level variance; we therefore read differences of a few millimeters, and all differences in collision counts, as directional (Supplementary Sec.~\ref{sec:suppl_stats}).

\subsection{Planning Performance}
\label{sec:exp_main}

\begin{table}[!tbp]
\centering
\caption{\textbf{nuScenes planning scores.} Average L2 (m) and collision rate (\%); $\diamond$: direct planner ego status. Baselines follow the compilation in ResWorld's Table~1~\cite{zhang2026resworld} (SSR$^\dagger$: its re-evaluation of the official model) and, for SpaceDrive+, its supplement; collision evaluators may differ across rows. Bold: lowest value; --: not reported.}
\label{tab:exp_main}
\fontsize{8.5}{10.2}\selectfont
\setlength{\tabcolsep}{1.7pt}
\renewcommand{\arraystretch}{1.12}
\begin{tabular*}{\linewidth}{@{\extracolsep{\fill}}lrrrr@{}}
\toprule
Method & \multicolumn{2}{c}{VAD-style} & \multicolumn{2}{c}{UniAD-style} \\
\cmidrule(lr){2-3}\cmidrule(lr){4-5}
 & L2 (m) $\downarrow$ & CR (\%) $\downarrow$ & L2 (m) $\downarrow$ & CR (\%) $\downarrow$ \\
\midrule
ST-P3~\cite{hu2022stp3} & 2.11 & 0.71 & 3.28 & 1.51 \\
UniAD~\cite{hu2023uniad} & 0.69 & 0.12 & 1.03 & 0.31 \\
VAD~\cite{jiang2023vad} & 0.72 & 0.22 & -- & -- \\
OccNet~\cite{sima2023occnet} & -- & -- & 2.14 & 0.72 \\
PARA-Drive~\cite{weng2024paradrive} & 0.48 & 0.25 & 0.83 & 0.30 \\
GenAD~\cite{zheng2024genad} & 0.52 & 0.19 & 0.91 & 0.43 \\
SSR$^\dagger$~\cite{li2025ssr} & 0.39 & 0.15 & 0.74 & 0.31 \\
SparseDrive-S~\cite{sun2025sparsedrive} & 0.61 & 0.08 & -- & -- \\
Drive-OccWorld~\cite{yang2025driveoccworld} & 0.47 & 0.11 & -- & -- \\
MomAD~\cite{song2025momad} & 0.60 & 0.09 & -- & -- \\
DiffusionDrive~\cite{liao2025diffusiondrive} & 0.57 & 0.08 & -- & -- \\
ResWorld~\cite{zhang2026resworld} & 0.35 & 0.07 & 0.65 & 0.23 \\
ResWorld$\diamond$~\cite{zhang2026resworld} & 0.30 & 0.06 & 0.59 & 0.17 \\
SpaceDrive+$\diamond$~\cite{li2026spacedrive} & 0.32 & 0.10 & 0.62 & 0.40 \\
\midrule
MOVER & 0.28 & 0.04 & 0.55 & 0.12 \\
MOVER$\diamond$ & \textbf{0.24} & \textbf{0.03} & \textbf{0.50} & \textbf{0.09} \\
\bottomrule
\end{tabular*}
\end{table}

\paragraph{Planning results.}
Without direct planner ego status, MOVER records average L2 errors of $0.2845\,\mathrm{m}$ and $0.5528\,\mathrm{m}$, with collision rates of $0.0445\%$ and $0.1172\%$ (Table~\ref{tab:exp_main}).
Relative to ResWorld's published scores, L2 falls by $19\%$ and $15\%$ and the UniAD-style collision rate from $0.23\%$ to $0.12\%$, the latter being $18$ endpoint collisions among $15{,}357$ observation--horizon pairs; MOVER also improves on ResWorld$\diamond$, which uses direct ego status, on all four metrics.
Because direct ego status can dominate open-loop nuScenes scores~\cite{li2024egostatus,zhai2023rethinking}, we treat this configuration, whose planner receives no ego-status vector while CAN-bus features still condition BEV encoding, as the primary result and use it for all nuScenes ablations.
The five added inference stages cost $12.1$ GFLOPs and $4.1\,\mathrm{ms}$ per frame on an RTX 3090 (Supplementary Table~\ref{tab:exp_runtime}); Supplementary Fig.~\ref{fig:exp_qualitative} shows qualitative plans.

\subsection{Closed-Loop Evaluation}
\label{sec:exp_closed_loop}

\begin{figure}[!t]
\centering
\includegraphics[width=\columnwidth]{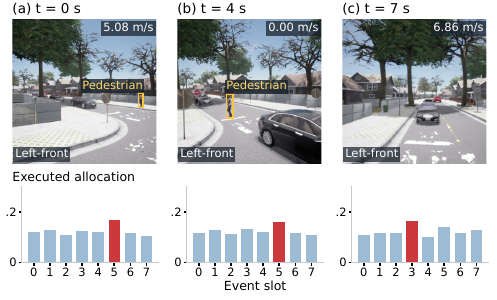}
\caption{\textbf{Approach, stop, and subsequent turn in CARLA.} Views at $0$, $4$, and $7$ seconds show ego speeds of $5.08$, $0.00$, and $6.86\,\mathrm{m/s}$ and their executed event allocations. Yellow boxes mark the recorded pedestrian.}
\label{fig:closed_loop_sequence}
\end{figure}

\begin{figure*}[!tbp]
\centering
\includegraphics[width=\linewidth]{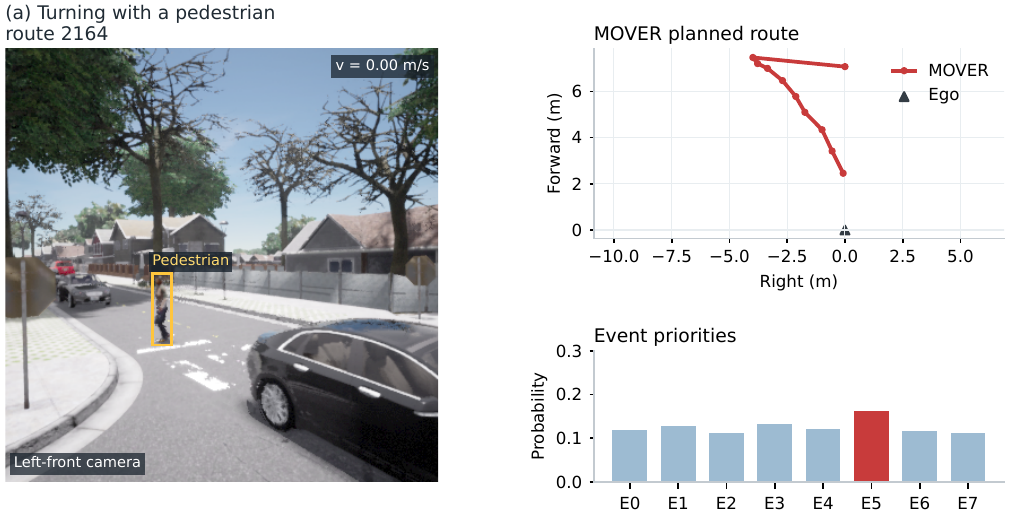}
\caption{\textbf{A turning scene with a pedestrian.} The left-front camera view is paired with the model's ten-point planned route in ego coordinates and its executed event allocation at the same observation.}
\label{fig:closed_loop_scene}
\end{figure*}

\begin{figure*}[!tbp]
\centering
\includegraphics[width=\linewidth]{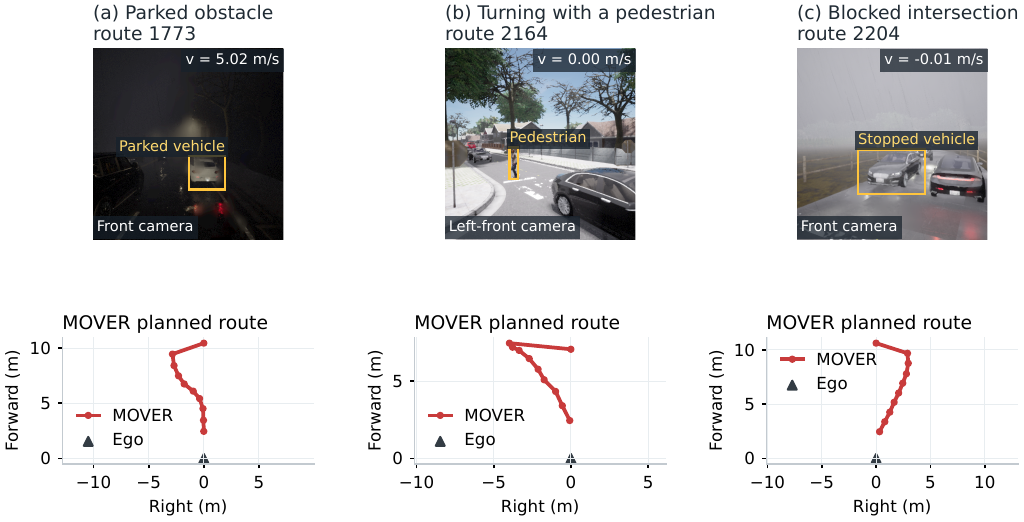}
\caption{\textbf{Three interactive driving scenarios.} Original camera views show a parked obstacle, a turn with a pedestrian, and a blocked intersection. Each is paired with the model's ten-point planned route at the same observation, expressed in ego coordinates.}
\label{fig:closed_loop_scenarios}
\end{figure*}

We further evaluate the event-routing interface on BridgeDrive--LEAD~\cite{liu2026bridgedrive,nguyen2026lead}, writing events into its BEV features and reading them through its planner queries, to test transfer to another base planner in closed loop.
Evaluation uses the original 220-route Bench2Drive benchmark~\cite{jia2024bench2drive}, CARLA 0.9.15, and the official scoring tools.
Table~\ref{tab:exp_closed_loop} reports a driving score of $96.61$ and a success rate of $90.00\%$ ($198/220$ routes) for seed~0.
Over three training seeds, MOVER reaches $96.52\pm0.65$ DS, against $95.92\pm0.40$ for uniform routing and $95.58\pm0.42$ for matched fine-tuning of the same base planner.
MOVER exceeds both controls in each paired seed, by $0.35$--$1.10$ DS over uniform routing, but three pairs do not make the mean gain of $0.60$ DS statistically significant (paired $95\%$ $t$-interval $[-0.47,1.67]$; Supplementary Table~\ref{tab:exp_closed_loop_paired}).
The published BridgeDrive--LEAD reports $96.34\pm0.55$ DS over the same protocol, while our unmodified base scores $95.35$ in one run, so MOVER exceeds the published mean by only $0.18$ DS, inside both seed spreads, and the paired gains are measured against controls below the published configuration.
Ability scores and infraction counts exist for seed~0 only: the study shows that the routing interface transfers without degrading a second planner, not a closed-loop safety gain.
Fig.~\ref{fig:closed_loop_sequence} shows an independent qualitative rollout; Supplementary Figs.~\ref{fig:closed_loop_scene} and~\ref{fig:closed_loop_scenarios} provide camera views with matched planned routes.

\begin{table*}[!t]
\centering
\caption{\textbf{Closed-loop performance on Bench2Drive (220 routes).} Baselines follow published evaluations~\cite{jia2024bench2drive,liu2026bridgedrive,nguyen2026lead,tang2025hipad}; Mean is recomputed from the five ability scores and can differ from published means by up to $0.06$. BridgeDrive--LEAD (published) is the base planner of our closed-loop study, reported as mean $\pm$ SD over three seeds~\cite{liu2026bridgedrive}. T2D: Think2Drive; $*$: expert feature distillation; temp/geo: temporal/geometric waypoints. $\dagger$: our controls on the BridgeDrive--LEAD base; $\ddagger$: mean of three paired training seeds (SDs in Sec.~\ref{sec:exp_closed_loop}). Mean averages the five ability scores. All metrics are higher-is-better; bold marks column maxima and -- an unavailable metric.}
\label{tab:exp_closed_loop}
\fontsize{8}{9.4}\selectfont
\setlength{\tabcolsep}{1pt}
\renewcommand{\arraystretch}{1.03}
\begin{tabular*}{\textwidth}{@{\extracolsep{\fill}}llrrrrrrrrrr@{}}
\toprule
Method & Expert & \multicolumn{4}{c}{Planning metrics} & \multicolumn{6}{c}{Driving abilities (\%)} \\
\cmidrule(lr){3-6}\cmidrule(lr){7-12}
 & & DS & SR (\%) & \shortstack{Efficiency\\(\%)} & \shortstack{Comfort\\(\%)} & Overtaking & Merging & \shortstack{Emergency\\brake} & \shortstack{Give\\way} & \shortstack{Traffic\\signs} & Mean \\
\midrule
VAD~\cite{jiang2023vad} & T2D & 42.35 & 15.00 & 157.94 & \textbf{46.01} & 24.44 & 8.11 & 18.64 & 20.00 & 19.15 & 18.07 \\
UniAD-Base~\cite{hu2023uniad} & T2D & 45.81 & 16.36 & 129.21 & 43.58 & 17.78 & 14.10 & 21.67 & 10.00 & 14.21 & 15.55 \\
TCP-traj$^*$~\cite{wu2022tcp} & T2D & 59.90 & 30.00 & 76.54 & 18.08 & 24.29 & 8.89 & 51.67 & 40.00 & 46.28 & 34.23 \\
DriveTransformer~\cite{jia2025drivetransformer} & T2D & 63.46 & 35.01 & 100.64 & 20.78 & 35.00 & 17.57 & 48.36 & 40.00 & 52.10 & 38.61 \\
ORION diffusion~\cite{fu2025orion} & T2D & 71.97 & 46.54 & -- & -- & -- & -- & -- & -- & -- & -- \\
ORION~\cite{fu2025orion} & T2D & 77.74 & 54.62 & 151.48 & 17.38 & 71.11 & 25.00 & 78.33 & 30.00 & 69.15 & 54.72 \\
HiP-AD~\cite{tang2025hipad} & T2D & 86.77 & 69.09 & 203.12 & 19.36 & 84.44 & 50.00 & 83.33 & 40.00 & 72.10 & 65.97 \\
\midrule
DiffusionDrive$_{\rm temp}$~\cite{liao2025diffusiondrive} & PDM-Lite & 77.68 & 52.72 & 248.18 & 24.56 & 26.67 & 50.63 & 68.33 & 50.00 & 76.32 & 54.39 \\
DiffusionDrive$_{\rm geo}$~\cite{liao2025diffusiondrive} & PDM-Lite & 80.79 & 58.18 & 245.34 & 15.49 & -- & -- & -- & -- & -- & -- \\
TransFuser++~\cite{zimmerlin2024datasetbiases} & PDM-Lite & 84.21 & 67.27 & -- & -- & 57.77 & 58.75 & 83.33 & 40.00 & 82.11 & 64.39 \\
SimLingo~\cite{renz2025simlingo} & PDM-Lite & 85.07 & 67.27 & \textbf{259.23} & 33.67 & 57.04 & 54.01 & 88.33 & \textbf{53.33} & 82.45 & 67.03 \\
BridgeDrive--PDM-Lite~\cite{liu2026bridgedrive} & PDM-Lite & 87.99 & 74.99 & 236.49 & 20.98 & 66.67 & 69.92 & 90.00 & 50.00 & 89.47 & 73.21 \\
\midrule
TFv6~\cite{nguyen2026lead} & LEAD & 95.20 & 86.80 & -- & -- & -- & -- & -- & -- & -- & -- \\
BridgeDrive--LEAD (published)~\cite{liu2026bridgedrive} & LEAD & 96.34 & 89.25 & 202.92 & 23.24 & 95.56 & 76.25 & \textbf{96.67} & 50.00 & \textbf{92.63} & 82.22 \\
Matched fine-tuning$^{\dagger\ddagger}$ & LEAD & 95.58 & 88.18 & 203.12 & 19.68 & -- & -- & -- & -- & -- & -- \\
Uniform routing$^{\dagger\ddagger}$ & LEAD & 95.92 & 88.64 & 205.24 & 19.52 & -- & -- & -- & -- & -- & -- \\
\midrule
MOVER (seed 0) & LEAD & \textbf{96.61} & \textbf{90.00} & 206.81 & 19.47 & \textbf{100.00} & \textbf{85.00} & 95.00 & 50.00 & 91.58 & \textbf{84.32} \\
MOVER$^{\ddagger}$ & LEAD & 96.52 & 89.85 & -- & -- & -- & -- & -- & -- & -- & -- \\
\bottomrule
\end{tabular*}
\end{table*}

\subsection{Ablation Studies}
\label{sec:exp_ablations}

\begin{table}[!t]
\centering
\caption{\textbf{Component ablation.} All rows retain the event branch. $R$: residual refiner; $P$: position loss; $C$: clearance loss; --: component not used. Entries are average L2 (m).}
\label{tab:exp_components}
\fontsize{8.5}{10.2}\selectfont
\setlength{\tabcolsep}{1.7pt}
\renewcommand{\arraystretch}{1.12}
\begin{tabular*}{\linewidth}{@{\extracolsep{\fill}}lcccrr@{}}
\toprule
ID & $R$ & $P$ & $C$ & \multicolumn{2}{c}{L2 (m) $\downarrow$} \\
\cmidrule(lr){5-6}
 &  &  &  & VAD & UniAD \\
\midrule
A0 & -- & -- & -- & 0.3137 & 0.6037 \\
A1 & $\checkmark$ & -- & -- & 0.2970 & 0.5752 \\
A2 & $\checkmark$ & $\checkmark$ & -- & 0.2894 & 0.5621 \\
A3 & $\checkmark$ & -- & $\checkmark$ & 0.2923 & 0.5649 \\
A4 & -- & $\checkmark$ & -- & 0.3012 & 0.5815 \\
A5 & $\checkmark$ & $\checkmark$ & $\checkmark$ & \textbf{0.2845} & \textbf{0.5528} \\
\bottomrule
\end{tabular*}
\end{table}

\begin{table}[!t]
\centering
\caption{\textbf{Routing and supervision controls} (non-$\diamond$), trained separately with shared capacity and refiner. (a)~Routing strategy under teacher KL; B2 is A5. (b)~Teacher target; S0 has no teacher. (c)~Routing without auxiliary losses: with B0/B2, a $2\times2$ routing$\times$loss design. --: not reported.}
\label{tab:exp_routes}
\fontsize{8.5}{10.2}\selectfont
\setlength{\tabcolsep}{1.2pt}
\renewcommand{\arraystretch}{1.12}
\begin{tabular*}{\linewidth}{@{\extracolsep{\fill}}llrrrr@{}}
\toprule
ID & Setting & \multicolumn{2}{c}{L2 (m) $\downarrow$} & \multicolumn{2}{c}{CR (\%) $\downarrow$} \\
\cmidrule(lr){3-4}\cmidrule(lr){5-6}
 &  & VAD & UniAD & VAD & UniAD \\
\midrule
\multicolumn{6}{@{}l}{\emph{(a) Model routing / read prior}} \\
B0 & Uniform / uniform & 0.3214 & 0.6285 & 0.0689 & 0.1888 \\
B1 & Learned / uniform & 0.3021 & 0.5842 & 0.0581 & 0.1563 \\
B2 & Learned / learned & \textbf{0.2845} & \textbf{0.5528} & \textbf{0.0445} & \textbf{0.1172} \\
\midrule
\multicolumn{6}{@{}l}{\emph{(b) Teacher target}} \\
S0 & None (task-trained) & 0.3111 & 0.6073 & 0.0621 & 0.1688 \\
S1 & Residual, $-z(d)$ & 0.3238 & 0.6314 & 0.0693 & 0.1912 \\
S2 & Sensitivity, $z(v)$ & 0.3021 & 0.5864 & 0.0550 & 0.1693 \\
B2 & Complementary & \textbf{0.2845} & \textbf{0.5528} & \textbf{0.0445} & \textbf{0.1172} \\
\midrule
\multicolumn{6}{@{}l}{\emph{(c) Model routing, no auxiliary losses}} \\
U1 & Uniform & 0.3061 & 0.5894 & -- & 0.2019 \\
A1 & Learned & 0.2970 & 0.5752 & -- & 0.1498 \\
\bottomrule
\end{tabular*}
\end{table}

\paragraph{Component effects.}
All rows in Table~\ref{tab:exp_components} retain the learned event branch, so the suite measures the refiner and the auxiliary losses given event context; A0 omits all three.
From A0 to A5, VAD- and UniAD-style L2 fall by $29.2$ and $50.9\,\mathrm{mm}$, and every available contrast lowers L2 (Supplementary Table~\ref{tab:exp_changes}).
The suite is not a full factorial and reports L2 only, so it does not isolate the effect of the clearance loss on collisions.

\paragraph{Routing effects.}
Table~\ref{tab:exp_routes}a compares configurations that all keep the router and its KL loss: B0 replaces the student route by a uniform distribution in both paths, B1 in the read prior only, and B2 uses it in both; from B0 to B2, L2 falls by $36.9$ and $75.7\,\mathrm{mm}$, of which the shared read prior (B1$\rightarrow$B2) accounts for $17.6$ and $31.4\,\mathrm{mm}$.
This contrast overstates the effect of routing alone, because routing interacts with the auxiliary losses (Table~\ref{tab:exp_routes}c): they lower UniAD-style L2 by $22.4\,\mathrm{mm}$ under learned routing (A1$\rightarrow$B2) but raise it by $39.1\,\mathrm{mm}$ under uniform routing (U1$\rightarrow$B0).
We therefore take the $14.2\,\mathrm{mm}$ gap without them (U1$\rightarrow$A1) as the conservative routing effect and $36.6\,\mathrm{mm}$ (U1 \vs B2) as MOVER's margin over the strongest uniform-routing model we trained.
Endpoint collisions total $29$ (B0), $31$ (U1), and $18$ (B2) among $15{,}357$ observation--horizon pairs, too few to resolve statistically (exact binomial $p=0.14$ and $0.09$); we report them as directional.

\paragraph{Router supervision.}
Table~\ref{tab:exp_routes}b changes only the teacher target: a router trained by the planning losses alone (S0) improves on uniform routing, but residual-only supervision (S1) is worse than S0 on all four metrics, and sensitivity-only supervision (S2) lowers L2 without lowering the UniAD-style collision rate.
The complementary target is best on all four metrics and reduces VAD- and UniAD-style L2 by $26.6$ and $54.5\,\mathrm{mm}$ relative to S0: neither signal suffices alone.
Supplementary Tables~\ref{tab:exp_route_loss_controls}--\ref{tab:exp_beta_sensitivity} add loss, event-count, and $\beta$ controls.

\subsection{Traffic Conditions and Routing Sensitivity}
\label{sec:exp_allocation_adds}

\paragraph{Traffic cohorts.}
Relative to task-trained routing (S0), MOVER lowers $3$-second L2 in every traffic cohort (Supplementary Table~\ref{tab:exp_condition_summary}): by $0.12$--$0.14\,\mathrm m$ ($11$--$14\%$) for waiting pedestrians, stopped junction vehicles, and partially occluded actors, by $0.17\,\mathrm m$ ($12\%$) for high-residual dynamic actors, and by $0.07\,\mathrm m$ ($7\%$) for low-residual background.
The gain is thus broad rather than specific to static or occluded actors: the cohorts do not show that MOVER selectively strengthens low-residual actors.
Consistently, the events covering the two static actors of Fig.~\ref{fig:novelty} receive route probabilities of $0.131$ and $0.117$, close to the uniform $0.125$ (Supplementary Table~\ref{tab:exp_static_actor_residual}).

\begin{figure*}[!tbp]
\centering
\includegraphics[width=0.9\linewidth]{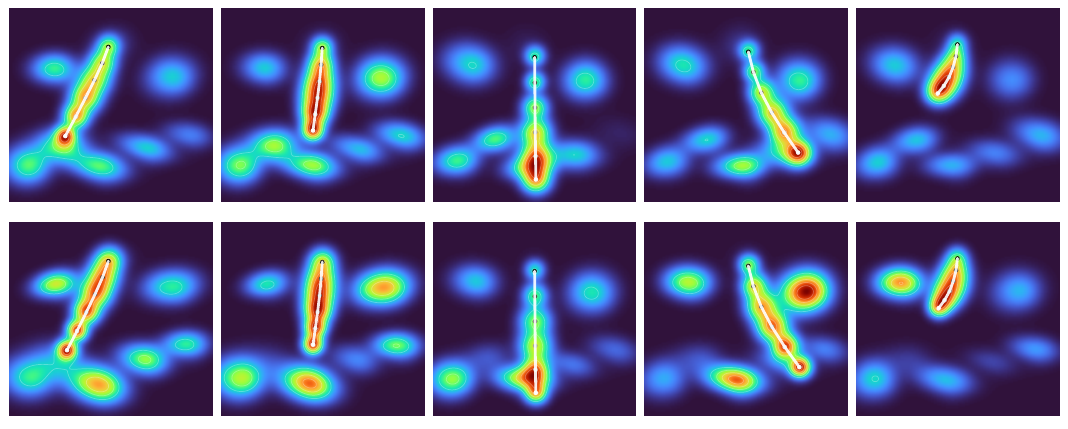}
\caption{\textbf{Spatial routing fields on nuScenes.} Five observations show smoothed spatial allocation rather than object-level attribution.}
\label{fig:exp_routing_maps}
\end{figure*}

\paragraph{Allocation and output sensitivity.}
Learned allocations shift with the navigation command, \eg toward events 6--7 for right turns and 0--1 for left turns (Supplementary Figs.~\ref{fig:exp_allocation} and~\ref{fig:exp_routing_maps}).
At a fixed checkpoint, replacing the student route with the teacher route shifts waypoints by only $1.6\,\mathrm{mm}$ on average, so the distilled router reproduces the teacher's effect; a uniform route shifts them by $8.2\,\mathrm{mm}$.
By the triangle inequality, the six-step mean L2 then changes by at most $8.2\,\mathrm{mm}$, well below the $36.9$--$75.7\,\mathrm{mm}$ between the separately trained B0 and B2.
Most of that gap therefore arises during training, where the allocation determines which event features are written and read: MOVER acts chiefly as a training-time routing prior, and we do not claim that scene-specific allocation at inference is necessary.

\section{Conclusion}
\label{sec:conclusion}

MOVER combines history sensitivity and pooled residual evidence to allocate supplementary BEV writing and trajectory reading.
On nuScenes it lowers L2 error relative to ResWorld, matched uniform-routing controls, and task-trained routers, and in closed loop it exceeds matched BridgeDrive--LEAD controls in every paired seed.
The allocation acts chiefly as a training-time routing prior.
Events are learned, overlapping spatial summaries rather than objects, so a route probability does not identify the actor behind a control action, and neither cohort results nor fixed-parameter route swaps establish object-specific causal relevance.
The teacher signals are model-relative: whole-frame deletion perturbs every event-history token, the two signals span different temporal windows, and neither is validated against counterfactual planning sensitivity.
nuScenes results lack seed-level variance, collision differences rest on $18$--$31$ events, and the routing gain depends on the auxiliary losses.
The closed-loop gain is not statistically significant, and NAVSIM~\cite{dauner2024navsim} evaluation remains future work.

{
    \small
    \bibliographystyle{ieeenat_fullname}
    \bibliography{main}
}

\clearpage
\maketitlesupplementary

\noindent Sections, equations, figures, and tables continue the numbering of the main paper, and numbered citations refer to its bibliography.

\section{Analysis and Algorithm}
\label{sec:suppl_analysis}

This section collects supporting analysis: a common form for MOVER's allocations and a stability bound for a fixed model.

\subsection{A Unifying Variational View}
\label{subsec:variational_view}

MOVER's core allocations share one principle.
For scores $s\in\mathbb R^n$, positive reference distribution $\pi$, and temperature $\tau>0$, define $J(p)=\langle p,s\rangle-\tau D_{\mathrm{KL}}(p\|\pi)$.
Consider the allocation problem,
\begin{equation}
	\max_{p\in\Delta_n}J(p).
	\label{eq:generic_variational}
\end{equation}

\begin{proposition}[KL-regularized allocation]
	\label{prop:variational}
	The unique maximizer of Eq.~\eqref{eq:generic_variational} is,
	\begin{equation}
		p_j^\star
		=
		\frac{\pi_j \exp(s_j/\tau)}
		{\sum_{k=1}^{n}\pi_k \exp(s_k/\tau)}.
		\label{eq:variational_solution}
	\end{equation}
	Moreover, for any $p\in\Delta_n$,
	\begin{equation}
		J(p^\star)-J(p)=\tau D_{\mathrm{KL}}(p\|p^\star).
		\label{eq:variational_gap}
	\end{equation}
\end{proposition}

\begin{proof}
Let $Z=\sum_k\pi_k\exp(s_k/\tau)$, so that $\log p_j^\star=\log\pi_j+s_j/\tau-\log Z$.
For any $p\in\Delta_n$, expanding the logarithm gives $\tau D_{\mathrm{KL}}(p\|p^\star)=\tau D_{\mathrm{KL}}(p\|\pi)-\langle p,s\rangle+\tau\log Z$, \ie, $J(p)=\tau\log Z-\tau D_{\mathrm{KL}}(p\|p^\star)$.
Since $J(p^\star)=\tau\log Z$, this is Eq.~\eqref{eq:variational_gap}, and because $D_{\mathrm{KL}}(p\|p^\star)=0$ only at $p=p^\star$, the maximizer is unique.
This is the Gibbs variational principle.
\end{proof}

This proposition recovers several pieces of MOVER as special cases: the event masks of Eq.~\eqref{eq:event_mask} use normalized spatial priors proportional to $\pi_j(u)$ and content scores $\ell_{ju}$; the teacher distribution of Eq.~\eqref{eq:comp_score} uses a uniform reference and complementary scores $c_j$; and route-conditioned trajectory reading uses content logits with a routing prior.
Spatial selection, event prioritization, and route-conditioned reading are instances of the same allocation principle, distinguished only by what plays the role of \emph{reference} and \emph{reward}.
The objective gap in Eq.~\eqref{eq:variational_gap} uses $D_{\mathrm{KL}}(p\|p^\star)$, while the distillation loss in Eq.~\eqref{eq:distill_loss} uses $D_{\mathrm{KL}}(q^T\|q^S)$.

\subsection{Pooled Residual}
\label{subsec:pooled_residual}

Let $r_u=X_{0u}-X_{1u}$ and $\bar r_j=\sum_u m_{ju}r_u$.
Since $\sum_u m_{ju}=1$, the squared pooled residual in Eq.~\eqref{eq:residual_evidence} obeys,
\begin{equation}
\sum_u m_{ju}\|r_u\|_2^2-d_j
=\sum_u m_{ju}\|r_u-\bar r_j\|_2^2
\geq 0.
\label{eq:residual_cancellation}
\end{equation}
The identity separates the squared pooled residual $d_j$ from variation among residual vectors within the event mask.

\subsection{Route Stability}
\label{subsec:stability}

The following bound relates teacher--student routing divergence to trajectory displacement for fixed model parameters.
Let $p,q\in\Delta_N$ be two route distributions.
Assume:
(i) the BEV write features are bounded,
(ii) the mappings from BEV to trajectory queries, from queries to base displacements, from route-conditioned event reading to trajectory context, and from refined context to residual displacement are Lipschitz on the region of interest.

Route mismatch then propagates through two paths: one through the BEV write of Eq.~\eqref{eq:bev_write}, which changes the future scene and therefore the trajectory queries, and one through the direct route-conditioned event reader of Eq.~\eqref{eq:traj_attention}.

\begin{theorem}[Two-path route stability]
	\label{thm:stability}
	There exists a constant $C_{\mathrm{route}}>0$ such that,
	\begin{equation}
		\|\widehat\delta(p)-\widehat\delta(q)\|_{\infty,2}
		\le
		C_{\mathrm{route}}\|p-q\|_1,
		\label{eq:step_stability}
	\end{equation}
	and consequently,
	\begin{equation}
		\|\widehat p_h(q^T)-\widehat p_h(q^S)\|_2
		\le
		h\,C_{\mathrm{route}}\sqrt{2D_{\mathrm{KL}}(q^T\|q^S)}.
		\label{eq:traj_stability}
	\end{equation}
\end{theorem}

\begin{proof}
Fix the input, event masks, event features, and model parameters while varying the route.
Let $\|f_j\|_2\le F_{\max}$ and $\|V_j\|_2\le V_{\max}$.
Assume the BEV-to-query map is $L_w$-Lipschitz from the BEV $\|\cdot\|_{\infty,2}$ norm to each query's Euclidean norm, and the content-logit vector is $L_\ell$-Lipschitz in that query with the $\ell_1$ norm.
Let the base decoder be $L_d$-Lipschitz, and let the residual decoder have separate constants $L_{gw}$ and $L_{gC}$ for its query and event-context inputs.

Since $0\le S_{uj}\le1$ and $r_j(p)-r_j(q)=\rho N(p_j-q_j)$, Eq.~\eqref{eq:bev_write} gives the explicit write bound,
\begin{equation}
\|\Delta B(p)-\Delta B(q)\|_{\infty,2}
\le W\|p-q\|_1,\qquad W=|\eta|\rho N F_{\max}.
\label{eq:write_route_bound}
\end{equation}
Consequently, $\|w_i(p)-w_i(q)\|_2\le L_wW\|p-q\|_1$.
The clamped log prior is $(k_\rho/\epsilon_p)$-Lipschitz in $\ell_1$ because $\epsilon_p>0$.
Softmax is $1$-Lipschitz from $\ell_1$ logits to $\ell_1$ probabilities, so Eq.~\eqref{eq:traj_attention} yields,
\begin{equation}
\|A_i(p)-A_i(q)\|_1
\le \left(L_\ell L_wW+\gamma k_\rho/\epsilon_p\right)\|p-q\|_1.
\label{eq:read_route_bound}
\end{equation}
The fixed event values then give $\|C_i(p)-C_i(q)\|_2\le V_{\max}\|A_i(p)-A_i(q)\|_1$.
Applying the decoder Lipschitz bounds to Eq.~\eqref{eq:delta_prediction} proves Eq.~\eqref{eq:step_stability} with,
\begin{equation}
C_{\mathrm{route}}
=(L_d+L_{gw})L_wW
+L_{gC}V_{\max}
\left(L_\ell L_wW+\gamma k_\rho/\epsilon_p\right).
\label{eq:route_constant}
\end{equation}
Summing the step differences over $h$ horizons and applying
$\|q^T-q^S\|_1\le\sqrt{2D_{\mathrm{KL}}(q^T\|q^S)}$
proves Eq.~\eqref{eq:traj_stability}.
\end{proof}

For a fixed predictor, the bound quantifies trajectory displacement from teacher--student route mismatch.

\paragraph{A data-dependent read constant.}
The factor $k_\rho/\epsilon_p$ in Eq.~\eqref{eq:route_constant} is a worst case that applies only when a routed probability reaches the clamp.
Let $q_{\min}=\min_j\max\{\epsilon_p/k_\rho,\min(p_j,q_j)\}$ denote the smallest routed probability of the two routes.
For $a,b\ge0$, $\lvert\log\max(k_\rho a,\epsilon_p)-\log\max(k_\rho b,\epsilon_p)\rvert\le\lvert a-b\rvert/\max\{\epsilon_p/k_\rho,\min(a,b)\}$, so the prior logits change by at most $\|p-q\|_\infty/q_{\min}\le\|p-q\|_1/(2q_{\min})$ in $\ell_\infty$, where the last step uses $\sum_j(p_j-q_j)=0$.
Softmax is also $1$-Lipschitz from $\ell_\infty$ logits to $\ell_1$ probabilities: at probabilities $s$, its Jacobian $J$ satisfies $\|Jx\|_1=\sum_j s_j\lvert x_j-\sum_k s_kx_k\rvert\le\|x\|_\infty$, a mean absolute deviation bounded by Popoviciu's inequality.
Equation~\eqref{eq:read_route_bound} therefore holds with $\gamma k_\rho/\epsilon_p$ replaced by $\gamma/(2q_{\min})$.
This factor never exceeds the original one and is far smaller whenever routed probabilities stay away from the clamp; for example, if every routed probability exceeds $0.01$, it is at most $50\gamma$ instead of about $10^6\gamma$.

\paragraph{Scope of the analysis.}
Proposition~\ref{prop:variational} is the classical Gibbs variational principle; we use it only to state the three allocations in one form.
Theorem~\ref{thm:stability} is a Lipschitz stability statement for fixed parameters: the plan varies at most linearly with the route, so the distillation error controls the gap between the teacher-routed and the student-routed plan.

\subsection{Training and Inference Algorithm}
\label{subsec:algorithm}

\begin{figure*}[t]
	\centering
	\includegraphics[width=\textwidth]{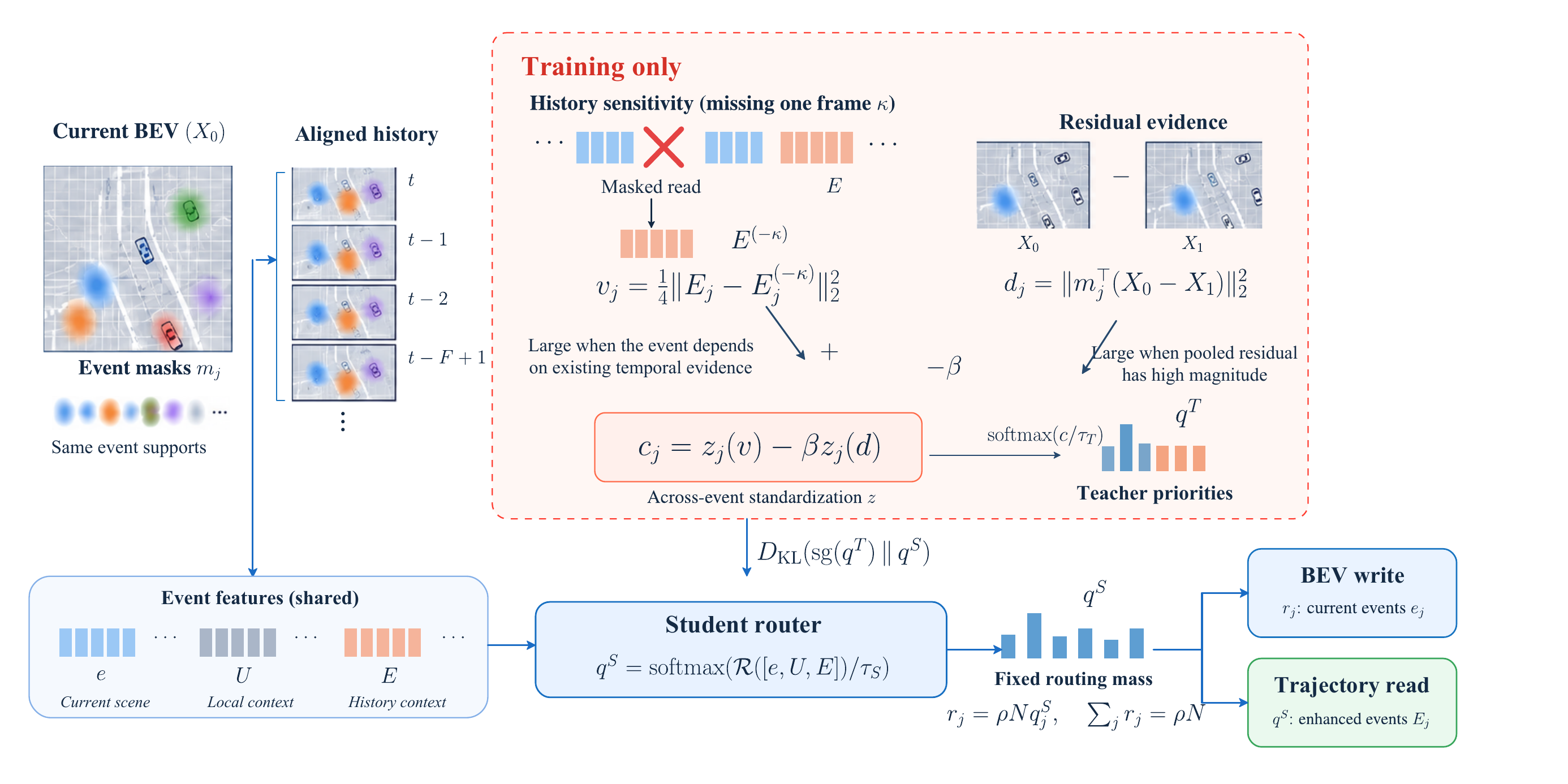}
	\caption{
		\textbf{The complementary routing module.}
		We discover current-frame event masks on the current BEV and reuse them to pool aligned history.
		A training-time teacher compares full-history and history-occluded event context, combines history sensitivity with residual evidence, and produces complementary priorities.
		The fixed-sum allocation writes a low-rank update into the future BEV and provides a prior for trajectory-conditioned event reading.
	}
	\label{fig:module}
\end{figure*}

Figure~\ref{fig:module} details the complementary routing module, and Algorithm~\ref{alg:mover} summarizes training and inference.

\begin{algorithm}[t]
	\caption{MOVER training and inference}
	\label{alg:mover}
	\begin{algorithmic}[1]
		\Require Ego-aligned BEV frames $\{X_f\}_{f=0}^{F-1}$, base predictor, route-weight scale $\rho$
		\State Extract current event tokens $\{e_j\}_{j=1}^{N}$ and masks $\{m_j\}_{j=1}^{N}$ from $X_0$
		\State Pool masked history and compute history-enhanced events $\{E_j\}_{j=1}^{N}$
		\If{training}
		\State Sample one historical frame $\kappa \sim \operatorname{Unif}\{1,\dots,F-1\}$
		\State Recompute perturbed event context $\{E_j^{(-\kappa)}\}_{j=1}^{N}$
		\State Compute $v_j$ by Eq.~\eqref{eq:history_sensitivity} and $d_j$ by Eq.~\eqref{eq:residual_evidence}
		\State Form teacher distribution $q^T$ by Eq.~\eqref{eq:comp_score}
		\EndIf
		\State Predict student routing distribution $q^S$ by Eq.~\eqref{eq:student_router}
		\State Convert $q^S$ to exact route mass $r_j=\rho N q_j^S$
		\State Write BEV correction $\Delta B(q)$ by Eq.~\eqref{eq:bev_write}
		\State Read route-conditioned event context $C_i(q)$
		\State Predict trajectory and, during training, optimize Eq.~\eqref{eq:full_objective}
	\end{algorithmic}
\end{algorithm}

\paragraph{Clearance geometry and annotation gate.}
For actor $a$ at horizon $h$, let $o_{bah}$ and $\theta_{bah}$ be its annotated future center and orientation, and let $\ell_{ba}$ and $w_{ba}$ be its box length and width.
We transform a candidate ego position into actor-local coordinates by,
\begin{equation}
z_{bah}(p)=R(-\theta_{bah})\bigl(p+(0,0.5)-o_{bah}\bigr).
\end{equation}
With margin $\mu=0.2\,\mathrm m$, the expanded half-extents are,
\begin{align}
h_x&=\ell_{ba}/2+0.925|\cos\theta_{bah}|+2.042|\sin\theta_{bah}|+\mu,\\
h_y&=w_{ba}/2+0.925|\sin\theta_{bah}|+2.042|\cos\theta_{bah}|+\mu.
\end{align}
Equation~\eqref{eq:clearance_loss} uses,
\begin{equation}
\psi_{bah}(p)=\max_{t\in\{x,y\}}
\frac{|z_{bah,t}(p)|}{\max(h_t,0.1)},
\end{equation}
and $\zeta_{bah}=a_{bah}\nu_{bh}\mathbf 1\{[1-\psi_{bah}(p^{\rm gt}_{bh})]_+^2<10^{-6}\}$, where $a_{bah}$ is the annotated actor-future validity flag.
The eligible actor set $\mathcal A_b$ contains annotation kind indices $2$--$8$ and $14$--$23$; an empty set contributes zero loss.

\paragraph{Implementation details.}
The reported configuration uses $N=8$, $\rho=0.25$, $\tau_T=\tau_S=0.5$, $\beta=1$, $\lambda_{\mathrm{KL}}=10^{-3}$, $\lambda_{\mathrm{pos}}=0.5$, and $\lambda_{\mathrm{clr}}=20$ with a $0.2$-m clearance margin.
The trajectory read uses $\gamma=1$ and $\epsilon_p=10^{-6}$; the trajectory residual is bounded componentwise by $\delta_{\max}=0.5\,\mathrm m$, and the learnable BEV scale $\eta$ starts at $0.1$.
The scale $\rho$ fixes total route mass; the BEV update norm also depends on the learned $\eta$, event features, and masks.
Because $\eta$ is initialized to $0.1$, the BEV write can still change the base decoder's input at initialization, whereas the direct trajectory residual starts at zero.
The final non-$\diamond$ routing comparison follows three localized-training epochs with four additional epochs under each routing strategy.
The AdamW base learning rate is $10^{-5}$ (weight decay $0.01$), with $10\times$ and $30\times$ learning-rate multipliers for the router and trajectory refiner, respectively.
In the non-$\diamond$ configuration, the planner sets \texttt{ego\_lcf\_feat\_idx=None}; CAN-bus features condition the BEV upstream.

\section{Endpoint Profile and Component Contrasts}
\label{sec:suppl_components}

\begin{figure}[t]
\centering
\includegraphics[width=\columnwidth]{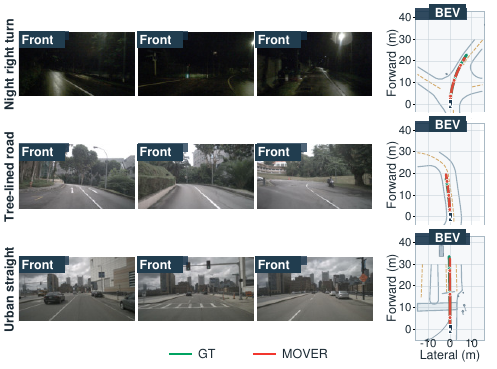}
\caption{\textbf{Qualitative planning on nuScenes.} Each row shows recorded front-camera frames and a BEV panel. Green denotes the recorded future trajectory and red the MOVER prediction.}
\label{fig:exp_qualitative}
\end{figure}

Figure~\ref{fig:exp_qualitative} shows matched camera observations and BEV plans.

\begin{table*}[t]
\centering
\caption{\textbf{Endpoint planning performance.} UniAD-style L2 (m) and collision rates (\%) by horizon. Published baselines follow the compilation in ResWorld's Table~1~\cite{zhang2026resworld}, except SpaceDrive+, which is taken from its supplementary Table~I~\cite{li2026spacedrive}. SSR$^\dagger$ is ResWorld's re-evaluation of the official SSR model; SSR itself reports $0.75\,\mathrm m$ and $0.15\%$ under this protocol~\cite{li2025ssr}. $\diamond$: direct planner ego status. The uniform and MOVER rows use matched local non-$\diamond$ evaluations.}
\label{tab:exp_endpoints}
\small
\setlength{\tabcolsep}{3.5pt}
\renewcommand{\arraystretch}{1.12}
\begin{tabular*}{\linewidth}{@{\extracolsep{\fill}}lrrrrrrrr@{}}
\toprule
Method & \multicolumn{4}{c}{L2 (m) $\downarrow$} & \multicolumn{4}{c}{Collision rate (\%) $\downarrow$} \\
\cmidrule(lr){2-5}\cmidrule(lr){6-9}
 & 1 s & 2 s & 3 s & Avg. & 1 s & 2 s & 3 s & Avg. \\
\midrule
ST-P3~\cite{hu2022stp3} & 1.72 & 3.26 & 4.86 & 3.28 & 0.44 & 1.08 & 3.01 & 1.51 \\
UniAD~\cite{hu2023uniad} & 0.48 & 0.96 & 1.65 & 1.03 & 0.05 & 0.17 & 0.71 & 0.31 \\
OccNet~\cite{sima2023occnet} & 1.29 & 2.13 & 2.99 & 2.14 & 0.21 & 0.59 & 1.37 & 0.72 \\
PARA-Drive~\cite{weng2024paradrive} & 0.40 & 0.77 & 1.31 & 0.83 & 0.07 & 0.25 & 0.60 & 0.30 \\
GenAD~\cite{zheng2024genad} & 0.36 & 0.83 & 1.55 & 0.91 & 0.06 & 0.23 & 1.00 & 0.43 \\
SSR$^\dagger$~\cite{li2025ssr} & 0.25 & 0.64 & 1.33 & 0.74 & 0.08 & 0.12 & 0.72 & 0.31 \\
ResWorld~\cite{zhang2026resworld} & 0.22 & 0.56 & 1.17 & 0.65 & 0.02 & 0.04 & 0.64 & 0.23 \\
ResWorld$\diamond$~\cite{zhang2026resworld} & 0.19 & 0.50 & 1.08 & 0.59 & 0.02 & 0.06 & 0.43 & 0.17 \\
SpaceDrive+$\diamond$~\cite{li2026spacedrive} & 0.20 & 0.53 & 1.13 & 0.62 & 0.10 & 0.31 & 0.80 & 0.40 \\
\midrule
MOVER (uniform) & 0.1894 & 0.5218 & 1.1743 & 0.6285 & \textbf{0.0195} & 0.0977 & 0.4493 & 0.1888 \\
MOVER & \textbf{0.1782} & \textbf{0.4615} & \textbf{1.0187} & \textbf{0.5528} & \textbf{0.0195} & \textbf{0.0586} & \textbf{0.2735} & \textbf{0.1172} \\
\bottomrule
\end{tabular*}
\end{table*}

\paragraph{Endpoint profiles.}
The uniform and MOVER endpoint rows use the non-$\diamond$ configuration; Table~\ref{tab:exp_main} gives the MOVER$\diamond$ averages.
Endpoint metrics use the $5{,}119$ observations with a complete future.
Their binary endpoint collision counts at $1$, $2$, and $3$ seconds are $(1,5,23)$ for uniform routing and $(1,3,14)$ for MOVER.
The corresponding average collision rates are $100(1+5+23)/(3\cdot5{,}119)=0.1888\%$ and $100(1+3+14)/(3\cdot5{,}119)=0.1172\%$.

\begin{figure}[t]
\centering
\includegraphics[width=\columnwidth]{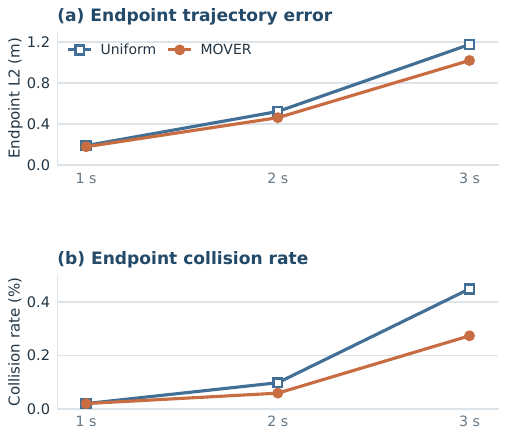}
\caption{\textbf{Planning accuracy across forecast horizons.}
The matched uniform-routing and MOVER models are evaluated on $5{,}119$ observations with a complete $3$-second future.
The panels display endpoint L2 and collision-rate summaries for the two routing strategies.}
\label{fig:exp_horizons}
\end{figure}

Figure~\ref{fig:exp_horizons} shows lower L2 at all three horizons and fewer collisions at $2$ and $3$ seconds for learned routing.

\begin{table}[t]
\centering
\caption{\textbf{Component L2 contrasts.} Each difference is destination minus source in millimeters; negative values indicate reductions. The last row combines $R$, $P$, and $C$.}
\label{tab:exp_changes}
\fontsize{8.5}{10.2}\selectfont
\setlength{\tabcolsep}{1.2pt}
\renewcommand{\arraystretch}{1.12}
\begin{tabular*}{\linewidth}{@{\extracolsep{\fill}}llrr@{}}
\toprule
Contrast & Change & \multicolumn{2}{c}{$\Delta$ L2 (mm)} \\
\cmidrule(lr){3-4}
 &  & VAD & UniAD \\
\midrule
A0$\rightarrow$A1 & \shortstack[l]{Add $R$;\\$P,C$ off} & $-16.7$ & $-28.5$ \\
A4$\rightarrow$A5 & \shortstack[l]{Add $R+C$;\\$P$ on} & $-16.7$ & $-28.7$ \\
A1$\rightarrow$A2 & \shortstack[l]{Add $P$;\\$C$ off} & $-7.6$ & $-13.1$ \\
A3$\rightarrow$A5 & \shortstack[l]{Add $P$;\\$C$ on} & $-7.8$ & $-12.1$ \\
A1$\rightarrow$A3 & \shortstack[l]{Add $C$;\\$P$ off} & $-4.7$ & $-10.3$ \\
A2$\rightarrow$A5 & \shortstack[l]{Add $C$;\\$P$ on} & $-4.9$ & $-9.3$ \\
\midrule
A0$\rightarrow$A5 & \shortstack[l]{Joint\\$R+P+C$} & $-29.2$ & $-50.9$ \\
\bottomrule
\end{tabular*}
\end{table}

\paragraph{Component contrasts.}
Table~\ref{tab:exp_components} lists the component configurations; Table~\ref{tab:exp_changes} gives their L2 differences.
A0 retains event writing; $R$, $P$, and $C$ vary on top of it.
The full $R+P+C$ configuration reduces L2 by $29.2\,\mathrm{mm}$ under the VAD-style protocol and $50.9\,\mathrm{mm}$ under the UniAD-style protocol relative to A0.

\section{Routing Controls and Output Sensitivity}
\label{sec:suppl_routing}

\begin{figure}[t]
\centering
\includegraphics[width=\columnwidth]{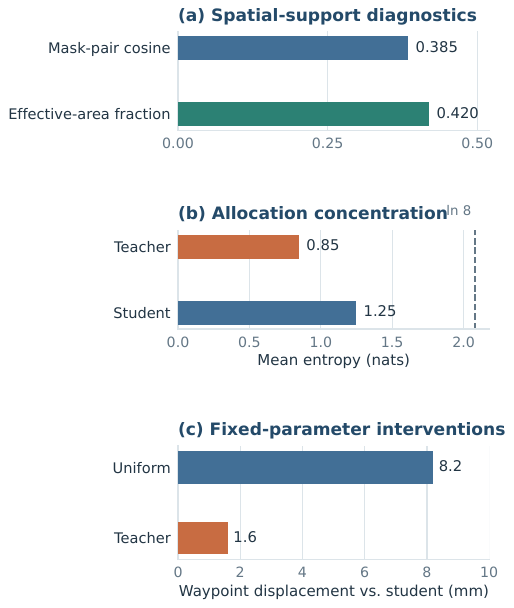}
\caption{\textbf{Localized route diagnosis.}
The panels show spatial-support statistics, mean teacher and student entropies, and mean Euclidean waypoint displacement from the unchanged student-route output.
Uniform and Teacher replace the student distribution in both routing paths at the same checkpoint; both overrides preserve total routing mass.}
\label{fig:exp_mechanism}
\end{figure}

\paragraph{Allocation and attention prior.}
Table~\ref{tab:exp_routes}a in the main paper compares B0 (uniform routing strategy), B1 (learned model routing with a uniform read prior), and B2 (learned routing and read prior).
The variants are trained separately with matched architecture, losses (including teacher--student KL), and optimizer settings.
In B0 and B1 the router is thus still trained by the KL term, but its output is replaced by the uniform distribution in both paths (B0) or in the read prior only (B1).
A uniform probability prior leaves the reader query-dependent because,
\begin{equation}
\begin{aligned}
&\operatorname{softmax}_j\!\left(
\ell_{ij}^{\mathrm{tr}}+\gamma\log(1/N)\right)\\
&\qquad=\operatorname{softmax}_j(\ell_{ij}^{\mathrm{tr}}).
\end{aligned}
\label{eq:exp_uniform_prior}
\end{equation}
The fixed-parameter Uniform and Teacher interventions in Fig.~\ref{fig:exp_mechanism} measure inference sensitivity of the localized model.

\paragraph{Fixed-parameter route sensitivity.}
Figure~\ref{fig:exp_mechanism} reports a mask-pair cosine of $0.385$ and an effective-area fraction of $0.420$ for the localized model.
Mask-pair cosine averages $\langle m_j,m_k\rangle/(\|m_j\|_2\|m_k\|_2)$ over distinct event pairs and observations.
The effective-area fraction averages $\exp(-\sum_u m_{ju}\log m_{ju})/M$ over events and observations, with $M$ BEV cells.
Mean teacher and student entropies are $0.85$ and $1.25$ nats, \ie, about $e^{0.85}\approx2.3$ and $e^{1.25}\approx3.5$ effective events of eight, so the student is more diffuse than its teacher.
Relative to student routing, the Uniform and Teacher interventions change waypoints by $8.2$ and $1.6\,\mathrm{mm}$ on average, respectively.
These distances quantify output sensitivity, and they bound the change in planning error: since $\bigl|\|\widehat p-p^{\mathrm{gt}}\|_2-\|\widehat p'-p^{\mathrm{gt}}\|_2\bigr|\le\|\widehat p-\widehat p'\|_2$, the uniform swap changes the six-step mean L2 on the same observations by at most $8.2\,\mathrm{mm}$, and the VAD- and UniAD-style averages by at most $15.0$ and $16.4\,\mathrm{mm}$ even if the entire shift fell on their most heavily weighted steps (weights $11/36$ and $1/3$ against the uniform $1/6$).
The inference-time allocation can therefore account for at most $41\%$ and $22\%$ of the $36.9$ and $75.7\,\mathrm{mm}$ that separate B0 from B2; the remainder reflects differences in the trained parameters.

\paragraph{Effect of event-value similarity.}
For a fixed trajectory query, let $a$ and $b$ be two attention distributions over event values $V_j$.
For any reference vector $V_\circ$,
\begin{equation}
C(a)-C(b)
=\sum_j(a_j-b_j)(V_j-V_\circ),
\label{eq:suppl_value_difference}
\end{equation}
since both distributions sum to one.
If the event values coincide, changing their weights leaves the readout unchanged.
An analogous limit holds for BEV writing: if all event masks coincide, their scatter weights are $S_{uj}=1/N$ wherever normalization is unclipped; if the projected write features also coincide at $f$, then,
\begin{equation}
\Delta B_u(q)
=\frac{\eta}{N}\sum_j\rho Nq_j f
=\eta\rho f,
\label{eq:exp_write_degeneracy}
\end{equation}
independently of the routing distribution.
Event-value diversity determines how strongly reallocation affects the readout.

\section{Population Structure and Paired Effects}
\label{sec:suppl_population}

\begin{figure}[t]
\centering
\includegraphics[width=\columnwidth]{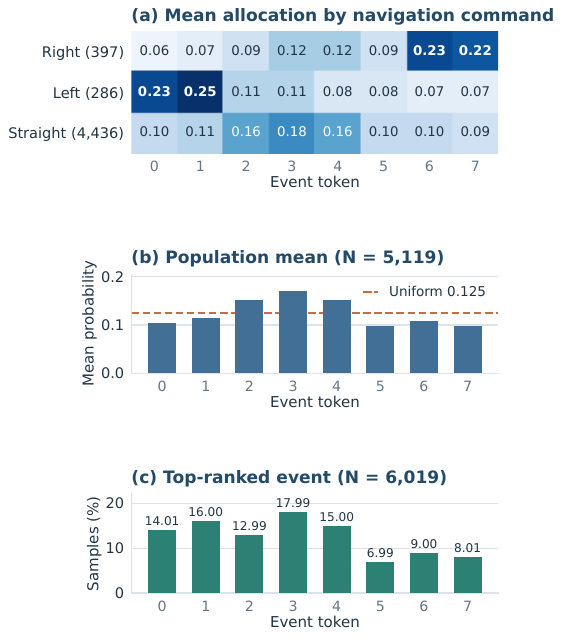}
\caption{\textbf{Learned event allocation.}
Command-conditioned means and their count-weighted population mean use $5{,}119$ complete-future observations; dominant-event counts use all $6{,}019$ validation observations. The dashed line marks the uniform reference $1/8$.}
\label{fig:exp_allocation}
\end{figure}

\paragraph{Allocation structure.}
Figure~\ref{fig:exp_allocation} shows command-dependent allocations: right turns emphasize events 6 and 7, left turns emphasize events 0 and 1, and straight driving emphasizes events 2--4.
The count-weighted mean over complete-future observations assigns event 3 probability $0.171$; across all validation observations, it is top-ranked in $1{,}083/6{,}019$ cases ($18\%$).
Because the allocation co-varies with the navigation command, part of it may act as a command-conditioned lateral prior rather than as scene-specific selection.

\begin{table}[t]
\centering
\caption{\textbf{Paired changes in six-step mean L2.} Each learned model is compared with an independently trained uniform-routing control. Paired scene-bootstrap intervals quantify scene variation for that model pair; negative changes favor learned routing.}
\label{tab:exp_paired}
\fontsize{8.5}{10.2}\selectfont
\setlength{\tabcolsep}{1.7pt}
\renewcommand{\arraystretch}{1.12}
\begin{tabular*}{\linewidth}{@{\extracolsep{\fill}}lrrl@{}}
\toprule
Supports & Epochs & Mean (mm) & 95\% CI (mm) \\
\midrule
Global & 1 & $-12.450$ & $[-18.230,-6.670]$ \\
Localized & 3 & $-24.810$ & $[-31.450,-18.170]$ \\
Localized & $3+4$ & $\mathbf{-38.620}$ & $\mathbf{[-46.850,-30.390]}$ \\
\bottomrule
\end{tabular*}
\end{table}

\paragraph{Paired full-path differences.}
Let $\varepsilon_b$ denote mean L2 over all six future steps for observation $b$.
The paired change $\Delta_b=1000(\varepsilon_b^{\mathrm{MOVER}}-\varepsilon_b^{\mathrm{Uniform}})$ is measured in millimeters, with negative values favoring MOVER.
The final localized comparison gives a mean change of $-38.620\,\mathrm{mm}$ and a $95\%$ interval of $[-46.850,-30.390]\,\mathrm{mm}$.
The paired statistic weights all six future steps equally; Table~\ref{tab:exp_routes} reports VAD-style prefix and UniAD-style endpoint averages.
The analysis resamples the $150$ validation scenes.

\raggedbottom
\setcounter{dbltopnumber}{4}
\setcounter{dblbotnumber}{2}
\setlength{\dbltextfloatsep}{10pt plus 2pt minus 2pt}
\setlength{\dblfloatsep}{10pt plus 2pt minus 2pt}
\renewcommand{\dblfloatpagefraction}{0.9}
\makeatletter
\setlength{\@fptop}{0pt}
\setlength{\@dblfptop}{0pt}
\setlength{\@fpsep}{12pt plus 2pt minus 2pt}
\setlength{\@dblfpsep}{12pt plus 2pt minus 2pt}
\makeatother
\section{Additional Routing Controls}
\label{sec:suppl_additional_controls}

\paragraph{Metric aggregation and collision counts.}
Let $e_{bt}=\|\widehat p_{bt}-p^{\mathrm{gt}}_{bt}\|_2$ be the position error of observation $b$ at future step $t$.
For horizon $h\in\{1,2,3\}$ seconds, let $T_h=2h$ and let $n_{\mathrm{val}}$ be the number of evaluated observations.
The two L2 summaries are,
\begin{equation}
\begin{aligned}
\mathrm{L2}_{\mathrm{pre}}(h)
 &= \frac{1}{n_{\mathrm{val}}}\sum_b
    \frac{1}{T_h}\sum_{t=1}^{T_h}e_{bt}, \\
\mathrm{L2}_{\mathrm{end}}(h)
 &= \frac{1}{n_{\mathrm{val}}}\sum_b e_{b,T_h},
\end{aligned}
\label{eq:exp_l2_protocols}
\end{equation}
and collision rates use the corresponding temporal aggregation of the planning evaluator's collision indicators.
Each reported average is the arithmetic mean of its $1$, $2$, and $3$-second scores, so the average VAD-style L2 weights the six future steps as,
\begin{equation}
\begin{aligned}
\mathrm{AvgL2}_{\mathrm{pre}}
=\big[&11(\bar e_1+\bar e_2)+5(\bar e_3+\bar e_4)\\
&{}+2(\bar e_5+\bar e_6)\big]/36,
\end{aligned}
\label{eq:exp_prefix_weights}
\end{equation}
for $\bar e_t=n_{\mathrm{val}}^{-1}\sum_b e_{bt}$, whereas the paired analysis uses the unweighted full-path mean.
The VAD-style columns average prefix metrics, whereas the UniAD-style columns average the $1$-, $2$-, and $3$-second endpoints (Eq.~\eqref{eq:exp_l2_protocols}).
For $n=5{,}119$ complete-future observations, $k_h$ counts endpoint collision positives at horizon $h$ and $K=\sum_{h\in\{1,2,3\}}k_h$; the UniAD-style average collision rate in percent is $100K/(3n)$.
An observation may contribute at multiple horizons, so $K$ counts sample--horizon positives.
The VAD-style prefix collision rate also requires intermediate-step indicators and cannot be recovered from $K$ alone.
One event changes the UniAD-style average rate by $100/(3n)=0.0065$ percentage points, and the matched contrasts of this paper involve between $17$ and $31$ events, so we quote counts alongside rates and do not interpret relative changes in the rate.

\begin{table*}[!tbp]
\centering
\caption{\textbf{Routing with auxiliary losses off and on.} $R$ denotes the trajectory refiner, $P$ the auxiliary position loss, and $C$ the clearance loss; all local runs use the non-$\diamond$ setting and retain the base trajectory objective. The ResWorld row is its published summary; local endpoint counts use $5{,}119$ complete-future observations. --: not available.}
\label{tab:exp_route_loss_controls}
\fontsize{8}{9.5}\selectfont
\setlength{\tabcolsep}{2.1pt}
\renewcommand{\arraystretch}{1.12}
\begin{tabular*}{\linewidth}{@{\extracolsep{\fill}}lccccrrcrr@{}}
\toprule
Configuration & Route & $R$ & $P$ & $C$ & VAD L2 (m) & UniAD L2 (m) & $(k_{1\mathrm s},k_{2\mathrm s},k_{3\mathrm s})$ & $K$ & UniAD Avg. CR (\%) \\
\midrule
ResWorld~\cite{zhang2026resworld} & -- & -- & -- & -- & 0.3500 & 0.6500 & -- & -- & 0.23 \\
U1 (uniform+$R$) & Uniform & $\checkmark$ & -- & -- & 0.3061 & 0.5894 & $(1,6,24)$ & 31 & 0.2019 \\
A1 ($R$ only) & Learned & $\checkmark$ & -- & -- & 0.2970 & 0.5752 & $(1,4,18)$ & 23 & 0.1498 \\
\midrule
B0 (full) & Uniform & $\checkmark$ & $\checkmark$ & $\checkmark$ & 0.3214 & 0.6285 & $(1,5,23)$ & 29 & 0.1888 \\
B2 (full) & Learned & $\checkmark$ & $\checkmark$ & $\checkmark$ & \textbf{0.2845} & \textbf{0.5528} & $(1,3,14)$ & 18 & \textbf{0.1172} \\
\bottomrule
\end{tabular*}
\end{table*}

\paragraph{Routing $\times$ auxiliary losses.}
U1 is trained independently from A1 with uniform routing; both omit the auxiliary position and clearance losses.
Together with B0 and B2, the four runs form a $2\times2$ design with the refiner always on (Table~\ref{tab:exp_routes}a,c).
Without the auxiliary losses, learned routing reduces VAD- and UniAD-style L2 by $9.1$ and $14.2\,\mathrm{mm}$ (U1$\rightarrow$A1) and the endpoint collision count from $31$ to $23$; with them, it reduces L2 by $36.9$ and $75.7\,\mathrm{mm}$ (B0$\rightarrow$B2) and the count from $29$ to $18$.
The two routing effects differ by $27.8$ and $61.5\,\mathrm{mm}$ because the auxiliary losses act in opposite directions: they lower L2 by $12.5$ and $22.4\,\mathrm{mm}$ under learned routing (A1$\rightarrow$B2) but raise it by $15.3$ and $39.1\,\mathrm{mm}$ under uniform routing (U1$\rightarrow$B0).
B0 is therefore not the strongest uniform-routing control; U1 is, and MOVER improves on it by $21.6$ and $36.6\,\mathrm{mm}$.
No cell has seed repeats, so we cannot separate this interaction from run-to-run variation, and we quote the $9.1$--$14.2\,\mathrm{mm}$ contrast as the conservative estimate of the routing effect.

\begin{table*}[!tbp]
\centering
\caption{\textbf{Event count and spatial prior.} All rows use the non-$\diamond$ setting. Collision rates average the three endpoints over $5{,}119$ observations. The latency increment is $\Delta t=t_{\rm variant}-t_{\rm ResWorld}$ on an RTX 3090 with PyTorch FP16 and batch size one. --: no spatial prior.}
\label{tab:exp_event_count}
\fontsize{8}{9.5}\selectfont
\setlength{\tabcolsep}{1.8pt}
\renewcommand{\arraystretch}{1.12}
\begin{tabular*}{\linewidth}{@{\extracolsep{\fill}}llrrcrrr@{}}
\toprule
Spatial prior & Grid & $N$ & UniAD L2 (m) & $(k_{1\mathrm s},k_{2\mathrm s},k_{3\mathrm s})$ & $K$ & Avg. CR (\%) & $\Delta t$ (ms) \\
\midrule
Gaussian grid & $1\times4$ & 4 & 0.5734 & $(1,4,17)$ & 22 & 0.1433 & +1.8 \\
Gaussian grid (default) & $2\times4$ & 8 & 0.5528 & $(1,3,14)$ & 18 & 0.1172 & +4.1 \\
Gaussian grid & $2\times8$ & 16 & \textbf{0.5491} & $(1,3,13)$ & 17 & \textbf{0.1107} & +8.7 \\
Gaussian grid & $4\times8$ & 32 & 0.5516 & $(1,3,14)$ & 18 & 0.1172 & +17.6 \\
Unanchored slots & -- & 8 & 0.5912 & $(1,5,19)$ & 25 & 0.1628 & +12.3 \\
\bottomrule
\end{tabular*}
\end{table*}

\paragraph{Event resolution.}
Among the tested event counts, $N=16$ gives the lowest UniAD-style L2 and collision rate, while the default $N=8$ uses less than half its reported additional latency.
The unanchored-slot variant has higher planning error than the spatially anchored configurations.

\begin{table*}[!tbp]
\centering
\caption{\textbf{Complementary-score coefficient.} The teacher score is $z(v)-\beta z(d)$. All rows use the non-$\diamond$ setting; endpoint counts and collision rates use $5{,}119$ complete-future observations.}
\label{tab:exp_beta_sensitivity}
\fontsize{8}{9.5}\selectfont
\setlength{\tabcolsep}{2.1pt}
\renewcommand{\arraystretch}{1.12}
\begin{tabular*}{\linewidth}{@{\extracolsep{\fill}}rrrrrr@{}}
\toprule
$\beta$ & VAD L2 (m) & UniAD L2 (m) & $(k_{1\mathrm s},k_{2\mathrm s},k_{3\mathrm s})$ & $K$ & UniAD Avg. CR (\%) \\
\midrule
0.0 & 0.3021 & 0.5864 & $(1,5,20)$ & 26 & 0.1693 \\
0.5 & 0.2912 & 0.5642 & $(1,4,16)$ & 21 & 0.1367 \\
1.0 & \textbf{0.2845} & \textbf{0.5528} & $(1,3,14)$ & 18 & \textbf{0.1172} \\
2.0 & 0.2974 & 0.5726 & $(1,4,18)$ & 23 & 0.1498 \\
\bottomrule
\end{tabular*}
\end{table*}

\paragraph{Complementary-score sensitivity.}
Removing the residual term ($\beta=0$) increases both L2 errors and collision rate relative to the default; $\beta=1$ is the best of the four evaluated settings on these metrics.

\section{Inference Profile}
\label{sec:suppl_inference_profile}

\begin{table*}[!tbp]
\centering
\caption{\textbf{Batch-one inference profile of ResWorld-based MOVER on nuScenes.} Six camera views, $200\times200$ BEV cells, $C=256$, $F=4$, and $N=8$. Times are means of $1{,}000$ model-forward passes after $50$ warm-ups; RTX cards use PyTorch FP16 and DRIVE Orin-X uses TensorRT FP16. The memory column is peak Orin allocation; --: not applicable. Stage rows decompose the cost added to ResWorld; relative increase is $100(\mathrm{MOVER}/\mathrm{ResWorld}-1)\%$ in each column. The training-only teacher is disabled.}
\label{tab:exp_runtime}
\fontsize{8}{9.5}\selectfont
\setlength{\tabcolsep}{2.1pt}
\renewcommand{\arraystretch}{1.12}
\begin{tabular*}{\linewidth}{@{\extracolsep{\fill}}lrrrrr@{}}
\toprule
Model or added stage & GFLOPs & RTX 3090 (ms) & RTX 4090 (ms) & Orin-X (ms) & Orin peak (MB) \\
\midrule
ResWorld & 284.2 & 38.4 & 24.6 & 62.5 & 1,450 \\
Current-event discovery & +0.8 & +0.3 & +0.2 & +0.6 & -- \\
Masked history pooling & +1.3 & +0.6 & +0.4 & +1.2 & -- \\
Event-to-history cross-attention & +4.6 & +1.2 & +0.7 & +2.4 & -- \\
Low-rank BEV scatter write & +1.8 & +0.8 & +0.5 & +1.5 & -- \\
Route-conditioned trajectory read & +3.6 & +1.2 & +0.8 & +2.3 & -- \\
\midrule
MOVER & 296.3 & 42.5 & 27.2 & 70.5 & 1,518 \\
Relative increase & +4.3\% & +10.7\% & +10.6\% & +12.8\% & +4.7\% \\
MOVER throughput (FPS) & -- & 23.5 & 36.8 & 14.2 & -- \\
\bottomrule
\end{tabular*}
\end{table*}

\paragraph{Asymptotic cost.}
With $M$ BEV cells, $F$ frames, $N$ events, $C$ channels, and $I$ trajectory queries, the added cost is $O(FMNC)$ for masked pooling, $O(FN^2C)$ for attention scores and value mixing, $O(MNC)$ for the write, and $O(INC)$ for the read.
For the small fixed $N$ used here, these terms scale linearly with BEV size and temporal length.

\paragraph{Measured cost.}
Relative to ResWorld, the five added inference stages total $12.1$ GFLOPs and add $4.1$, $2.6$, and $8.0\,\mathrm{ms}$ per frame on the RTX 3090, RTX 4090, and DRIVE Orin-X, respectively.
The complete model-forward latency on Orin-X is $70.5\,\mathrm{ms}$ ($14.2$ FPS); the teacher perturbation is used only during training.

\section{Closed-loop Qualitative Examples}
\label{sec:suppl_closed_loop_qualitative}

The following camera views come from independent CARLA 0.9.15 rollouts of the model evaluated in Table~\ref{tab:exp_closed_loop}.
Yellow rectangles project recorded actor boxes into the camera view; event indices denote model slots.

\section{Traffic Context and Event Routing}
\label{sec:suppl_traffic_context}

\begin{figure*}[!tbp]
\centering
\includegraphics[width=\linewidth]{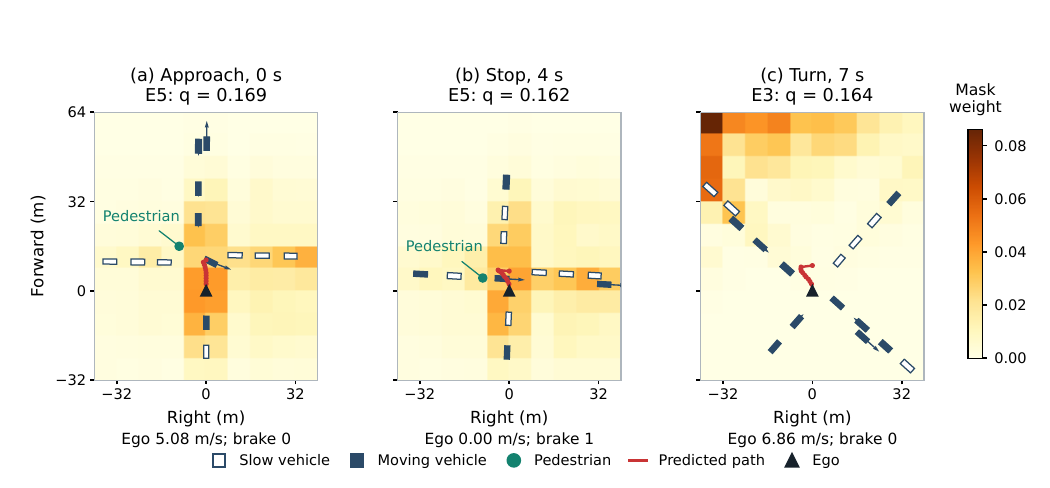}
\caption{\textbf{Spatial support of the dominant events in Figure~\ref{fig:closed_loop_sequence}.} Color shows the top-ranked event's normalized mask on the native $10\times12$ feature grid, with a shared scale. We place recorded actor boxes and predicted paths in nominal BEV coordinates ($8\,\mathrm m$ per cell). Hollow vehicle boxes denote speeds at most $0.2\,\mathrm{m/s}$; arrows show one second of recorded velocity. These masks pool spatial context and have overlapping support.}
\label{fig:traffic_context_support}
\end{figure*}

\paragraph{Approach and stop.}
The pedestrian in Figure~\ref{fig:closed_loop_sequence} is stationary at $t=0$ and moves at $1.80\,\mathrm{m/s}$ at $t=4\,\mathrm s$.
Its ego-relative position changes from $(16.0,-9.6)$ to $(4.6,-9.5)\,\mathrm m$ in forward/right coordinates.
Event 5 receives the largest allocation at both observations: $0.169$ and $0.162$, compared with the uniform value $0.125$.
Its broad support covers traffic around the junction, including moving and slow vehicles.
The controller applies full braking at $t=4\,\mathrm s$, when ego speed reaches zero.

\paragraph{Global allocation and local support.}
At the pedestrian position, event 1 has the largest bilinearly sampled mask value at both observations, although event 5 has the largest global route probability.
The two masks overlap; a slot pools evidence from a region rather than representing an object class.
On this $10\times12$ grid one cell spans about $8\,\mathrm m$, so a pedestrian occupies a small fraction of a single cell and no event mask can localize it; the closed-loop allocations describe coarse regions of the scene.
The planner also combines the route prior with query-dependent content when reading events (Eq.~\eqref{eq:traj_attention}).
Thus, a high route probability alone does not identify the actor that determined a control action.

\paragraph{Subsequent turn.}
At $t=7\,\mathrm s$, event 3 becomes dominant: its probability increases from $0.132$ at $t=4\,\mathrm s$ to $0.164$, while event 5 decreases from $0.162$ to $0.142$.
Event 3 places $62.7\%$ of its mask mass in the forward-left region $x\in[32,64]$, $y\in[-40,0]\,\mathrm m$, which contains a recorded stopped vehicle.
Ego travels at $6.86\,\mathrm{m/s}$ with the brake released and a left-steering command of $-0.599$.
The sequence shows a change in the spatial context receiving the largest allocation as ego turns.

\paragraph{Planning effects.}
The recorded controls describe the executed behavior; the fixed-parameter nuScenes interventions test sensitivity to allocation.
Replacing the learned route with a uniform route shifts waypoints by $8.2\,\mathrm{mm}$ on average (Figure~\ref{fig:exp_mechanism}).
These interventions support an effect on planning outputs without assigning the CARLA braking action to one event.

\section{Additional Event Diagnostics}
\label{sec:suppl_event_diagnostics}

\paragraph{Stationarity and pooled residual.}
Table~\ref{tab:exp_static_actor_residual} compares the two CARLA observations in Fig.~\ref{fig:novelty}.
For each actor, we select the event with the largest bilinearly sampled mask at its recorded center.
Both actors are stationary, but the selected events have residual ranks $2/8$ and $6/8$ within their respective observations.
Actor speed therefore does not determine the pooled feature residual; event support includes surrounding traffic and shared context.
Nor is either selected event prioritized: their route probabilities are $0.1305$ and $0.1166$ against the uniform $0.125$, and the stopped vehicle's event falls below uniform while carrying one of the larger residuals.
The two observations therefore illustrate the setting that motivates MOVER; they are not evidence that MOVER up-weights stationary actors, and the route probability of an overlapping, region-level event should not be read as object-level attention.
Establishing an actor-specific effect requires paired counterfactuals (inserting, removing, displacing, or re-timing one actor) that track the event residual, the sensitivity, the route weight, the plan, and the clearance together; we leave them to future work.

\begin{table*}[!tbp]
\centering
\caption{\textbf{Recorded static actors and their event support.} $d_j$ is the squared norm of the CARLA adapter's pooled latent-feature residual; rank 1 denotes the smallest residual among eight events. The uniform routing probability is $0.125$. Event indices are zero-based.}
\label{tab:exp_static_actor_residual}
\fontsize{8.5}{10.2}\selectfont
\setlength{\tabcolsep}{3pt}
\renewcommand{\arraystretch}{1.15}
\begin{tabular*}{\linewidth}{@{\extracolsep{\fill}}llrrrr@{}}
\toprule
Actor & Speed (m/s) & Event & $d_j$ & Residual rank & $q_j$ \\
\midrule
Waiting pedestrian & $0$ & 1 & $2.3575\times10^{-4}$ & $2/8$ & 0.1305 \\
Stopped junction vehicle & $<0.01$ & 1 & $4.3746\times10^{-5}$ & $6/8$ & 0.1166 \\
\bottomrule
\end{tabular*}
\end{table*}

\section{Additional Scenario and Supervision Evaluations}
\label{sec:suppl_scenario_supervision}

We evaluate nuScenes planning by interaction condition and maneuver using complete-future observations and $3$-second endpoint L2.
Condition groups may overlap.
The task-trained router omits teacher KL while retaining the refiner and auxiliary losses (S0 in Table~\ref{tab:exp_routes}; Table~\ref{tab:exp_supervision_controls}).

\begin{table*}[!tbp]
\centering
\caption{\textbf{Planning across interaction conditions.} Endpoint L2 at $3$ seconds, in meters, on the same $n$ complete-future observations per condition; conditions may overlap. $\Delta$ is the change of MOVER relative to the task-trained router, in meters and in percent.}
\label{tab:exp_condition_summary}
\fontsize{8.5}{10}\selectfont
\setlength{\tabcolsep}{3pt}
\renewcommand{\arraystretch}{1.08}
\begin{tabular*}{\linewidth}{@{\extracolsep{\fill}}lrrrrrr@{}}
\toprule
Condition & $n$ & ResWorld & Task-trained router & MOVER & $\Delta$ (m) & $\Delta$ (\%) \\
\midrule
Waiting pedestrian & 612 & 0.92 & 0.88 & \textbf{0.76} & $-0.12$ & $-14$ \\
Stopped vehicle at junction & 487 & 1.05 & 1.01 & \textbf{0.89} & $-0.12$ & $-12$ \\
Partially occluded actor & 1,386 & 1.28 & 1.23 & \textbf{1.09} & $-0.14$ & $-11$ \\
High-residual dynamic actor & 538 & 1.46 & 1.40 & \textbf{1.23} & $-0.17$ & $-12$ \\
Low-residual background & 947 & 1.01 & 0.99 & \textbf{0.92} & $-0.07$ & $-7$ \\
\bottomrule
\end{tabular*}
\end{table*}

\begin{table*}[!tbp]
\centering
\caption{\textbf{Maneuver-specific planning at $3$ seconds.} Endpoint L2 in meters on the complete-future observations of each group; $\Delta$ as in Table~\ref{tab:exp_condition_summary}. The three navigation commands partition the $5{,}119$ observations; the stop-and-go and braking/yielding groups may overlap with them and with each other.}
\label{tab:exp_maneuver_summary}
\fontsize{8.5}{10}\selectfont
\setlength{\tabcolsep}{3pt}
\renewcommand{\arraystretch}{1.08}
\begin{tabular*}{\linewidth}{@{\extracolsep{\fill}}lrrrrrr@{}}
\toprule
Maneuver & $n$ & ResWorld & Task-trained router & MOVER & $\Delta$ (m) & $\Delta$ (\%) \\
\midrule
Straight & 4,436 & 1.12 & 1.08 & \textbf{0.97} & $-0.11$ & $-10$ \\
Left turn & 286 & 1.58 & 1.52 & \textbf{1.38} & $-0.14$ & $-9$ \\
Right turn & 397 & 1.45 & 1.40 & \textbf{1.27} & $-0.13$ & $-9$ \\
Stop-and-go & 418 & 0.95 & 0.91 & \textbf{0.79} & $-0.12$ & $-13$ \\
Braking and yielding & 653 & 0.88 & 0.84 & \textbf{0.74} & $-0.10$ & $-12$ \\
\bottomrule
\end{tabular*}
\end{table*}

\paragraph{Scenario profile.}
MOVER reduces L2 relative to task-trained routing by $0.07$--$0.17\,\mathrm m$ across interaction conditions and $0.10$--$0.14\,\mathrm m$ across maneuvers.
Left and right turns remain harder than straight driving ($1.38$, $1.27$, and $0.97\,\mathrm m$).
The relative reductions are similar for static, occluded, and dynamic cohorts ($11$--$14\%$) and across maneuvers ($9$--$13\%$), and the largest absolute reduction is for high-residual dynamic actors.
The scenario tables therefore indicate a broad improvement of the planning representation; they do not show that the gain is concentrated on stationary or low-residual actors.

\begin{table*}[!tbp]
\centering
\caption{\textbf{Routing-supervision controls on nuScenes.} Adapted models share the ResWorld initialization; routing variants share the refiner, auxiliary losses, freezing policy, and training budget. ResWorld is the published reference; the remaining rows are local trained models.}
\label{tab:exp_supervision_controls}
\fontsize{8.5}{10}\selectfont
\setlength{\tabcolsep}{2.2pt}
\renewcommand{\arraystretch}{1.12}
\begin{tabular*}{\linewidth}{@{\extracolsep{\fill}}llrrrrr@{}}
\toprule
Method & Teacher target & $\lambda_{\mathrm{KL}}$ & VAD L2 (m) $\downarrow$ & VAD CR (\%) $\downarrow$ & UniAD L2 (m) $\downarrow$ & UniAD CR (\%) $\downarrow$ \\
\midrule
ResWorld (published) & -- & -- & 0.35 & 0.07 & 0.65 & 0.23 \\
ResWorld + fine-tuning & -- & 0 & 0.3432 & 0.0694 & 0.6421 & 0.2210 \\
Task-trained routing & -- & 0 & 0.3111 & 0.0621 & 0.6073 & 0.1688 \\
Sensitivity teacher & $z(v)$ & 0.001 & 0.3021 & 0.0550 & 0.5864 & 0.1693 \\
Residual teacher & $-z(d)$ & 0.001 & 0.3238 & 0.0693 & 0.6314 & 0.1912 \\
MOVER & $z(v)-z(d)$ & 0.001 & \textbf{0.2845} & \textbf{0.0445} & \textbf{0.5528} & \textbf{0.1172} \\
\bottomrule
\end{tabular*}
\end{table*}

\paragraph{Trained-model contrasts.}
Table~\ref{tab:exp_routes}b in the main paper summarizes the routing variants of this table.
MOVER reduces VAD- and UniAD-style L2 by $58.7$ and $89.3\,\mathrm{mm}$ relative to fine-tuned ResWorld, and by $26.6$ and $54.5\,\mathrm{mm}$ relative to task-trained routing.
The sensitivity-only teacher is the $\beta=0$ setting of Table~\ref{tab:exp_beta_sensitivity}; residual-only supervision underperforms task-trained routing on all four metrics.
For reference, A0 of Table~\ref{tab:exp_components}, which adds the event branch without the refiner or the auxiliary losses, is $29.5$ and $38.4\,\mathrm{mm}$ below fine-tuned ResWorld; a model that adds the refiner and the auxiliary losses without event context is not among the controls.

\section{Additional Closed-loop Controls and Variability}
\label{sec:suppl_closed_loop_controls}

\begin{table*}[!tbp]
\centering
\caption{\textbf{Local Bench2Drive controls under LEAD expert supervision.} Each evaluation covers 220 routes. Multi-seed entries are mean $\pm$ standard deviation. MOVER uses seed 0, as in Table~\ref{tab:exp_closed_loop}; Table~\ref{tab:exp_closed_loop_seeds} gives its three-seed results.}
\label{tab:exp_closed_loop_controls}
\fontsize{8.5}{10}\selectfont
\setlength{\tabcolsep}{3pt}
\renewcommand{\arraystretch}{1.12}
\begin{tabular*}{\linewidth}{@{\extracolsep{\fill}}lrrrrr@{}}
\toprule
Method & Seeds & DS $\uparrow$ & SR (\%) $\uparrow$ & Efficiency (\%) $\uparrow$ & Comfort (\%) $\uparrow$ \\
\midrule
ResWorld (CARLA adaptation) & 3 & $88.73\pm1.15$ & $77.27\pm1.20$ & $185.42\pm5.83$ & $21.83\pm1.52$ \\
BridgeDrive--LEAD & 1 & 95.35 & 87.73 & 201.35 & 19.85 \\
BridgeDrive--LEAD + matched fine-tuning & 3 & $95.58\pm0.42$ & $88.18\pm0.79$ & $203.12\pm3.45$ & $19.68\pm0.71$ \\
BridgeDrive--LEAD + uniform routing & 3 & $95.92\pm0.40$ & $88.64\pm0.46$ & $205.24\pm2.89$ & $19.52\pm0.58$ \\
MOVER (seed 0) & 1 & 96.61 & 90.00 & 206.81 & 19.47 \\
\bottomrule
\end{tabular*}
\end{table*}

\begin{table*}[!tbp]
\centering
\caption{\textbf{Closed-loop variability across training seeds.} Configurations are fixed, simulator conditions are paired across methods, and each seed covers 220 routes. Summary rows give mean $\pm$ standard deviation.}
\label{tab:exp_closed_loop_seeds}
\fontsize{8.5}{10}\selectfont
\setlength{\tabcolsep}{3pt}
\renewcommand{\arraystretch}{1.08}
\begin{tabular*}{\linewidth}{@{\extracolsep{\fill}}llrr@{}}
\toprule
Method & Training seed & DS $\uparrow$ & SR (\%) $\uparrow$ \\
\midrule
ResWorld (CARLA adaptation) & 0 & 89.52 & 78.18 \\
 & 1 & 87.41 & 75.91 \\
 & 2 & 89.25 & 77.73 \\
 & Mean $\pm$ SD & $88.73\pm1.15$ & $77.27\pm1.20$ \\
\midrule
BridgeDrive--LEAD + matched fine-tuning & 0 & 95.95 & 88.64 \\
 & 1 & 95.12 & 87.27 \\
 & 2 & 95.68 & 88.64 \\
 & Mean $\pm$ SD & $95.58\pm0.42$ & $88.18\pm0.79$ \\
\midrule
BridgeDrive--LEAD + uniform routing & 0 & 96.25 & 89.09 \\
 & 1 & 95.48 & 88.18 \\
 & 2 & 96.02 & 88.64 \\
 & Mean $\pm$ SD & $95.92\pm0.40$ & $88.64\pm0.46$ \\
\midrule
MOVER & 0 & 96.61 & 90.00 \\
 & 1 & 95.83 & 88.64 \\
 & 2 & 97.12 & 90.91 \\
 & Mean $\pm$ SD & $\mathbf{96.52\pm0.65}$ & $\mathbf{89.85\pm1.14}$ \\
\bottomrule
\end{tabular*}
\end{table*}

\begin{table*}[!tbp]
\centering
\caption{\textbf{Failures and infractions on Bench2Drive.} Counts are affected routes among 220 evaluations of MOVER seed 0; categories may overlap. Runtime terminations are separate from driving infractions.}
\label{tab:exp_closed_loop_infractions}
\fontsize{8.5}{10}\selectfont
\setlength{\tabcolsep}{3pt}
\renewcommand{\arraystretch}{1.08}
\begin{tabular*}{\linewidth}{@{\extracolsep{\fill}}lrr@{}}
\toprule
Failure / infraction & Routes $\downarrow$ & Rate (\%) $\downarrow$ \\
\midrule
Unsuccessful route & 22 & 10.00 \\
Pedestrian collision & 1 & 0.45 \\
Vehicle collision & 9 & 4.09 \\
Static-object collision & 0 & 0.00 \\
Red-light violation & 2 & 0.91 \\
Stop-sign violation & 0 & 0.00 \\
Outside-route lane violation & 1 & 0.45 \\
Route-deviation record & 3 & 1.36 \\
Route timeout & 0 & 0.00 \\
Scenario timeout & 0 & 0.00 \\
Vehicle blocked & 0 & 0.00 \\
Emergency-vehicle yielding violation & 5 & 2.27 \\
Runtime termination & 2 & 0.91 \\
\bottomrule
\end{tabular*}
\end{table*}

\paragraph{Variability and remaining failures.}
Across three seeds, MOVER improves mean DS by $0.60$ and success rate by $1.21$ percentage points over the three-seed uniform-routing control.
For MOVER seed 0, $22$ routes are unsuccessful; vehicle collisions ($9$ routes), emergency-vehicle yielding violations ($5$), and runtime terminations ($2$) remain.
Infraction counts and ability scores are available for MOVER seed~0 only and have no counterpart for the controls, so the closed-loop study does not compare safety events between methods.
The CARLA-adapted ResWorld of Table~\ref{tab:exp_closed_loop_controls} is a reference for the base predictor of the nuScenes study; MOVER is not evaluated on it, so the closed-loop study tests the routing interface on a different base planner.

\begin{table*}[!tbp]
\centering
\caption{\textbf{Paired closed-loop differences across training seeds.} MOVER minus control for the seed-matched runs of Table~\ref{tab:exp_closed_loop_seeds}; success-rate differences are also given in routes of 220. Intervals are two-sided $95\%$ paired $t$-intervals with two degrees of freedom.}
\label{tab:exp_closed_loop_paired}
\fontsize{8.5}{10}\selectfont
\setlength{\tabcolsep}{3pt}
\renewcommand{\arraystretch}{1.08}
\begin{tabular*}{\linewidth}{@{\extracolsep{\fill}}llrrrrr@{}}
\toprule
Control & Metric & Seed 0 & Seed 1 & Seed 2 & Mean $\pm$ SD & $95\%$ interval \\
\midrule
Uniform routing & DS & $+0.36$ & $+0.35$ & $+1.10$ & $0.60\pm0.43$ & $[-0.47,\,1.67]$ \\
 & SR (pp) & $+0.91$ & $+0.46$ & $+2.27$ & $1.21\pm0.94$ & $[-1.13,\,3.55]$ \\
 & SR (routes) & $+2$ & $+1$ & $+5$ & -- & -- \\
\midrule
Matched fine-tuning & DS & $+0.66$ & $+0.71$ & $+1.44$ & $0.94\pm0.44$ & $[-0.15,\,2.02]$ \\
 & SR (pp) & $+1.36$ & $+1.37$ & $+2.27$ & $1.67\pm0.52$ & $[0.37,\,2.97]$ \\
 & SR (routes) & $+3$ & $+3$ & $+5$ & -- & -- \\
\bottomrule
\end{tabular*}
\end{table*}

\paragraph{Paired seed differences.}
Table~\ref{tab:exp_closed_loop_paired} pairs the runs by training seed.
MOVER is ahead of both controls in all three pairs, by one to five additional successful routes of $220$, but with three pairs only the success-rate gain over matched fine-tuning has an interval that excludes zero, and the smallest two-sided sign-test $p$-value attainable with three pairs is $0.25$.
The intervals assume approximately normal differences, which three values cannot verify.
We therefore describe the closed-loop result as a gain that is consistent in sign but not statistically resolved.

\section{Statistical Notes and Open Controls}
\label{sec:suppl_stats}

\paragraph{Seed variance.}
The nuScenes tables report one value per configuration and no variance across training seeds, in contrast to the three-seed closed-loop study.
Differences of a few millimeters between separately trained nuScenes models, such as the $2.2\,\mathrm{mm}$ between B1 and S2 in UniAD-style L2, should therefore not be interpreted.

\paragraph{Collision counts.}
The matched local contrasts involve $K=29$ (B0), $31$ (U1), $23$ (A1), and $18$ (B2) endpoint collisions among $3n=15{,}357$ observation--horizon pairs (Table~\ref{tab:exp_route_loss_controls}).
Treating the events as independent, exact conditional binomial tests give $p=0.14$ for B2 against B0 and $p=0.09$ for B2 against U1.
Independence is optimistic here, because one observation can collide at several horizons and consecutive observations share a scene, so the true evidence is weaker still.
Collision differences between matched local models are therefore not statistically resolved, and we report them as directional.


\end{document}